\documentclass[sigconf]{acmart}
\AtBeginDocument{%
  \providecommand\BibTeX{{%
    \normalfont B\kern-0.5em{\scshape i\kern-0.25em b}\kern-0.8em\TeX}}}

\usepackage{multirow}
\usepackage{enumitem}
\usepackage{xspace}
\usepackage{makecell}

\newcommand{\modelname}{\textit{GraphAdapter}\xspace}

\copyrightyear{2024}
\acmYear{2024}
\setcopyright{acmlicensed}\acmConference[WWW '24]{Proceedings of the ACM
Web Conference 2024}{May 13--17, 2024}{Singapore, Singapore}
\acmBooktitle{Proceedings of the ACM Web Conference 2024 (WWW '24), May
13--17, 2024, Singapore, Singapore}
\acmDOI{10.1145/3589334.3645627}
\acmISBN{979-8-4007-0171-9/24/05}

\begin{document}

\title{Can GNN be Good Adapter for LLMs? }

\author{Xuanwen Huang}
\email{xwhuang@zju.edu.cn}
\orcid{0000-0002-4668-4570}
\affiliation{%
  \institution{Zhejiang University}
  \city{Hangzhou}
  \country{China}
}

\author{Kaiqiao Han}
\email{kaiqiaohan@zju.edu.cn}
\affiliation{%
  \institution{Zhejiang University}
  \city{Hangzhou}
  \country{China}
}
\author{Yang Yang}
\authornotemark[2]
\email{yangya@zju.edu.cn}
\affiliation{%
  \institution{Zhejiang University}
  \city{Hangzhou}
  \country{China}
}

\author{Dezheng Bao}
\email{baodezheng@zju.edu.cn}
\affiliation{%
  \institution{Zhejiang University}
  \city{Hangzhou}
  \country{China}
}

\author{Quanjin Tao}
\email{taoquanjin@zju.edu.cn}
\affiliation{%
  \institution{Zhejiang University}
  \city{Hangzhou}
  \country{China}
}

\author{Ziwei Chai}
\email{zwchai@zju.edu.cn}
\affiliation{%
  \institution{Zhejiang University}
  \city{Hangzhou}
  \country{China}
}

\author{Qi Zhu}
\authornotemark[1]
\email{qzhuamazon@amazon.com}
\affiliation{%
  \institution{Amazon Web Services}
  \city{Santa Clara}
  \country{USA}
}
\renewcommand{\shortauthors}{Xuanwen Huang, et al.}
\renewcommand{\thefootnote}{\fnsymbol{footnote}}
\begin{abstract}
Recently, large language models (LLMs) have demonstrated superior capabilities in understanding and zero-shot learning on textual data, promising significant advances for many text-related domains. In the graph domain, various real-world scenarios also involve textual data, where tasks and node features can be described by text. These text-attributed graphs (TAGs) have broad applications in social media, recommendation systems, etc. Thus, this paper explores how to utilize LLMs to model TAGs. 
Previous methods for TAG modeling are based on million-scale LMs. When scaled up to billion-scale LLMs, they face huge challenges in computational costs. Additionally, they also ignore the zero-shot inference capabilities of LLMs. 
Therefore, we propose \modelname, which uses a graph neural network (GNN) as an efficient adapter in collaboration with LLMs to tackle TAGs. 
In terms of efficiency, the GNN adapter introduces only a few trainable parameters and can be trained with low computation costs. The entire framework is trained using auto-regression on node text (next token prediction). Once trained, \modelname can be seamlessly fine-tuned with task-specific prompts for various downstream tasks.
Through extensive experiments across multiple real-world TAGs, \modelname based on Llama 2 gains an average improvement of approximately 5\% in terms of node classification. Furthermore, GraphAdapter can also adapt to other language models, including RoBERTa, GPT-2. The promising results demonstrate that GNNs can serve as effective adapters for LLMs in TAG modeling.
\footnotetext[2]{Yang Yang is the corresponding author.}
\footnotetext[1]{This work was done before the author joined Amazon.}
\footnotetext{The code is available at: \url{https://github.com/zjunet/GraphAdapter}}
\end{abstract}

\begin{CCSXML}
<ccs2012>
<concept>
<concept_id>10010147.10010178</concept_id>
<concept_desc>Computing methodologies~Artificial intelligence</concept_desc>
<concept_significance>300</concept_significance>
</concept>
<concept>
<concept_id>10002951.10003227.10003351</concept_id>
<concept_desc>Information systems~Data mining</concept_desc>
<concept_significance>300</concept_significance>
</concept>
</ccs2012>
\end{CCSXML}

\ccsdesc[300]{Computing methodologies~Artificial intelligence}
\ccsdesc[300]{Information systems~Data mining}
\keywords{Graph Neural Networks, Large Language Model, Text-Attributed Graph}

\maketitle
\renewcommand{\thefootnote}{\arabic{footnote}}

\section{Introduction}
Graphs are ubiquitous in the real world \cite{berge2001theory}. 
In the past, graph structures have been extensively explored and utilized in many machine learning applications \cite{newman2002random, ying2018graph}.
In many practical cases, the nodes in graphs have textual features, known as Textual-Attributed Graphs (TAGs) \cite{yang2021graphformers}.  
For example, in social media \cite{kim2020multimodal}, nodes represent users and node features are user profiles. Nodes in TAGs have both textual and structural data, which both reflect their intrinsic properties. 
Combining textual and structural data to modeling TAGs is an exciting new exploration for both graph machine learning and language modeling, which can benefit the application of graphs.

In TAGs, a complex correlation exists between the structural and textual data of nodes. Understanding this correlation can benefit the modeling of TAGs \cite{corley2010text}. 
In Figure 1, user ``Bob'' frequently browses daily news on social media, as evidenced by the descriptions in his user profile. Users similar to Bob, who have many followers and often browse news nodes, are also likely interested in news. In other words, a graph can supplement textual attributes on a node through structural proximity. 
Graph Neural Networks (GNNs) are the de facto machine learning models for leveraging textual information alongside graph structures in TAGs.
However, there's a lack of a unified GNN architecture compatible with different language models, especially the powerful foundation models.

Recently, there has been a surge in studies investigating effective ways to model both textual and structural data in TAGs. Some of these studies emphasize optimizing a cascading architecture that combines GNNs and LMs (\textbf{cascading GNN-LMs}) \cite{yang2021graphformers, zhao2022learning}. One major challenge with these models is the extreme amount of additional computational cost brought by the message-passing mechanism. To this end, several studies have successfully reduced the memory and computational overheads of such cascaded models by freezing partial or full parameters of the backbone language models \cite{liu2019fine, li2021adsgnn}. 
Large language models exhibit superior multi-task and few-shot learning capabilities across various real-world applications \cite{brown2020language}. However, when considering cascading GNN-LMs, existing techniques cannot be scaled up to billion-scale models like Llama 2 \cite{touvron2023llama}. Another pioneering research has ventured to fine-tune language models using unsupervised graph information (\textbf{self-supervised GNN-LMs}) \cite{chien2021node, mavromatis2023train}. 
For instance, GIANT \cite{chien2021node} fine-tunes language models through a neighbor prediction task, subsequently using the refined language model to extract node representations for downstream tasks. 
% \hxw{These studies have conclusively shown that graphs can indeed aid language models in comprehending textual information.} 
In these methods, PLMs can indirectly incorporate graph information during the tuning process thereby enhancing their capability to process TAGs.
However, they separate the training of GNNs and LMs, potentially leading to sub-optimal graph-aware tuning results.

\begin{figure}[t!]
	\centering
	\includegraphics[width=0.4\textwidth]{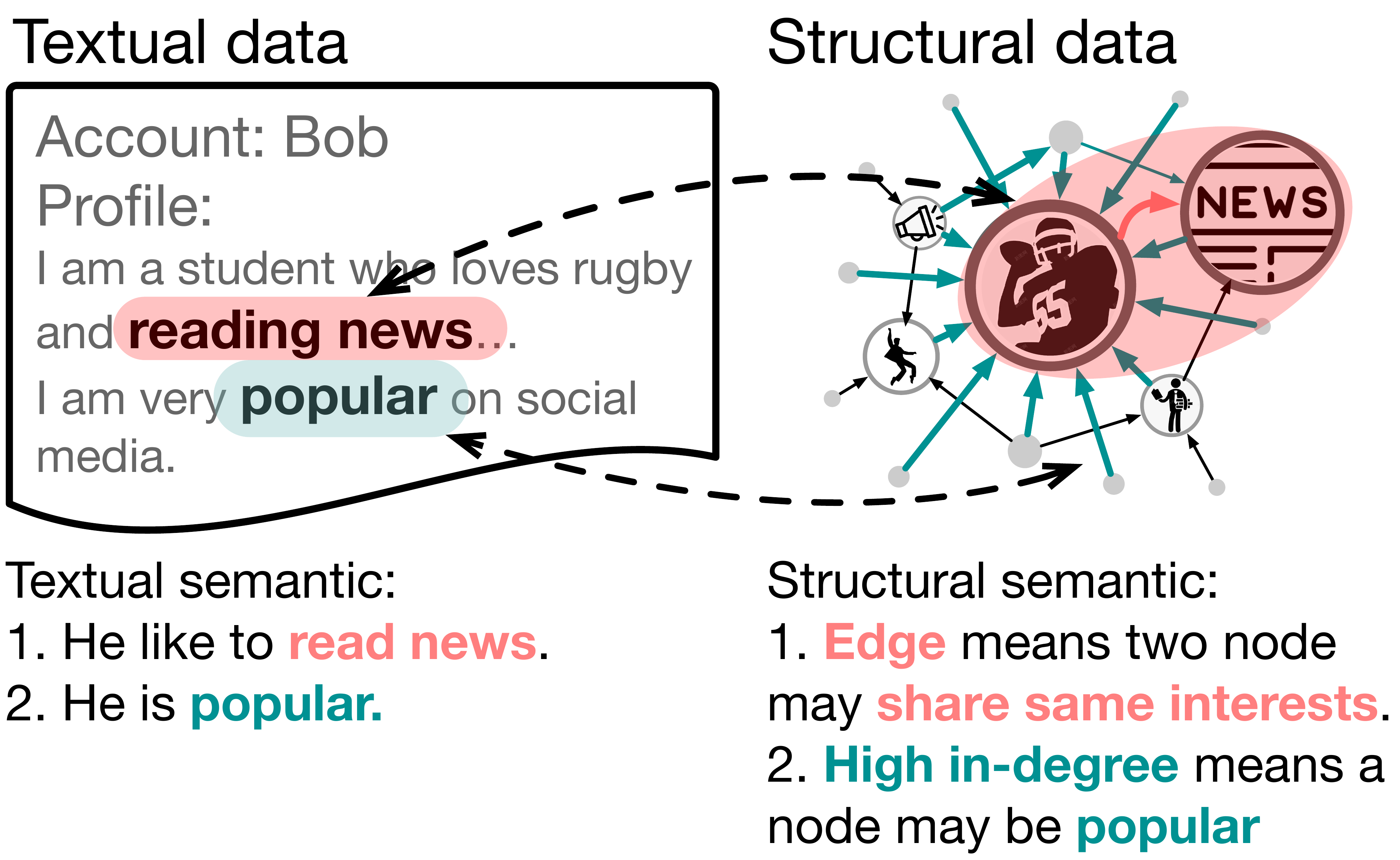}
	\caption{An example of the correlation existing in the structural and textual data of nodes in social networks. 
	\normalsize
	}
	\label{fig:intro}
\end{figure}

Instead of using graph information as supervision, we believe graph structure can enrich textual features through language modeling. 
In our previous example, structural proximity can be used to infer the user's preference even if he or she does not mention it in the profile.
So, unlike self-supervised methods, we consider pre-training a framework that can combine graph-aware structure and LLMs by leveraging rich textual features. 
However, traditional frameworks like cascading GNNs and LLMs face efficiency issues in pre-training scenarios.
Therefore, inspired by works on parameter-efficient tuning of LLMs \cite{hu2021lora, li2021prefix, liu2021p}, we propose the use of GNNs as efficient adapters for LLMs (i.e., \modelname). 
In \modelname, the LM is frozen and the final output of the LM is altered by the trainable adapter GNNs. 
\modelname offers several advantages:
\begin{itemize}
\item \textbf{Lightweight:} A GNN adapter introduces a few trainable parameters and low computational costs.
\item \textbf{Language-aware graph pre-training}: Using language to supervise the modeling of graph structure, which can help LLMs comprehend both textual and structural information.
\item \textbf{Convenient tuning:} Once a graph-specific adapter is pre-trained, it can be fine-tuned for multiple downstream tasks.
\end{itemize}

Now we present the details of \modelname with respect to pre-training and fine-tuning of the adapter GNNs. To capture the data distribution of the graph, we employ parameter-efficient tuning of LLMs on node texts. This approach is similar to the continual training of language models \cite{sun2020ernie} except GNN is the tuning parameter, which helps reduce the distribution discrepancy between the pre-training corpus and target data. 
To further enhance efficiency, we utilize the GNN adapter exclusively at the transformer's final layer. 
It ensures that all transformer computational processes are executed just once and then can be cached for adapter training.
Besides, we perform mean-pooling on the predicted logits from a GNN adapter and LLMs then optimize their final results of the next-word prediction, which can help adapters focus more on the graph-related tokens. 
Once the adapter is trained, one can use \modelname together with the backbone LLMs on various downstream tasks. For instance, we use a classification head atop the embeddings of the last token to fine-tune for node classification. 

To verify the effectiveness of \modelname, we conduct extensive experiments on multiple real-world TAGs including social and citation networks.
\modelname achieves an improvement of 4.7\% over state-of-the-art cascaded GNN-LM methods and 5.4\% over self-supervised GNN-LMs on average, with 30X fewer training parameters and storage.
Moreover, once \modelname is pre-trained, it can be conveniently fine-tuned for various tasks. Our ablation analysis shows that the pre-training step consistently improves the model performance across different graphs.
We summarize our contributions as follows,
\begin{itemize}[leftmargin=*,topsep=1pt]
    \item \modelname is a novel approach that harnesses the large language models on graph structure data with parameter-efficient tuning.
    \item We propose a residual learning procedure to pre-train the GNN adapter with the LLMs. The pre-training step significantly improves the fine-tuning performance of \modelname.
    \item We conduct extensive experiments on large-scale TAGs using state-of-the-art open-sourced large language models (GPT-2 1.5B \cite{radford2018improving} and Llama 2 13B \cite{touvron2023llama}). The results demonstrate that \textit{Graph-Adapter} can also reap the benefits of a larger model.
\end{itemize}

\section{Related work}
Modeling text-attributed graphs has attracted much attention in academia, which requires modeling both textual and structural data.

\textbf{Modeling semantics and graph structure}. 
Understanding the semantics is a key part of modeling TAG. 
With the advent of Transformers \cite{vaswani2017attention}, pre-trained language models have made breakthrough progress in modeling semantics \cite{devlin2018bert,zhou2019semantic}. 
These methods leverage massive unlabeled text through unsupervised methods like auto-regressive \cite{lester2021power} and auto-encoding pre-training \cite{liu2019roberta, he2020deberta} to pre-train Transformers and then fine-tune to downstream tasks. Since language models have a large number of parameters, fine-tuning efficiency is low and requires numerous training samples. Therefore, some work proposed using adapter modules to reduce the number of fine-tuning parameters. For example, LoRA \cite{hu2021lora} trains a sparse matrix appended to the original parameters while keeping the language model frozen. Some work proposed using prompts to directly adapt language models to downstream tasks without fine-tuning. 
Furthermore, some work proposed prompt tuning \cite{jiang2022promptbert,liu2021p}, which adds a trainable prompt and only trains the added prompt during training, greatly reducing the number of parameters. Another aspect of modeling TAG is modeling the structural information. With the proposal of GNNs \cite{hamilton2017inductive}, modeling graph structure achieved remarkable success. Many works \cite{xu2018powerful, li2021training} have explored GNN architectures extensively, and these methods have achieved breakthrough progress in graph structure modeling.

\textbf{Modeling TAGs}. 
However, despite the success of language models and GNNs in their respective areas, how to utilize them to model text-attributed graphs still has many challenges. \textbf{(1) Cascading GNN-LM}: Directly cascading these two models is straightforward but has limitations, mainly high computational overhead. 
Since GNNs are mostly based on message-passing, they need to compute representations for many nodes simultaneously. 
Using language models to model so many text features requires huge memory and time costs. To address this, some work \cite{liu2019fine} proposed freezing the language model to reduce the computation needed for cascading. 
Some work \cite{li2021adsgnn, jin2023patton} proposed neighbor sampling but that reduces the graph information captured.
Therefore, recently some work tried joint training of LMs and GNNs through knowledge distillation \cite{mavromatis2023train} or Expectation Maximization algorithms \cite{zhao2022learning}.
\textbf{(2) Self-supervised GNN-LMs}: some methods \cite{chien2021node, mavromatis2023train} directly supervise language model fine-tuning through graph-related tasks, to help language models better understand the textual information in text-attributed graphs. The language model is then combined with GNNs by freezing the language model. This approach demonstrates the inherent connections between graph structure and text in TAGs. However, current research in this direction has limitations in that the LM and graph are separate, and the language model cannot directly perceive graph information. It also does not utilize the inherent connections between language and graphs to help GNNs better learn structural features. 
\textbf{(3) LLMs for Graph}: With the breakthrough progress made by LLMs on textual tasks \cite{touvron2023llama, zeng2022glm}, recently many works have emerged exploring how to directly utilize LLMs to understand text-attributed graphs \cite{chen2023exploring}. For example, by converting the graph to text \cite{guo2023gpt4graph, yuan2023evaluating}, or by converting it to a graph representation as part of a prompt \cite{tian2023graph}. Some works also explored using large models to enhance the textual features of text-attributed graphs \cite{he2023explanations, duan2023simteg}. However, this paper is more focused on how to leverage the semantic information in text-attributed graphs to help us model text-attributed graphs. Therefore, this type of method not be further elaborated.
\section{Background}
Before introducing the proposed method, it's important to understand some basic concepts and the background of pre-trained language models, graph neural networks, and text-attributed graphs.
\subsection{Pretrained Language Model}

\noindent \textbf{Textual data.} Textual data can be formulated as $\mathbb{D} =\{d_1,d_2...d_K\}$. It can be tokenized into a sequence of tokens $\mathbb{S} = \{s_1, s_2, ... , s_L\}$, where $s_i$ represents a specific token-id. In most cases, the first token in the sentence (i.e., $s_{0}$) is $[\mathbf{CLS}]$, indicating the beginning of this sentence.

\noindent\textbf{Framework of PLMs}. A PLM consists of a multi-layer transformer encoder that takes a sentence $\mathbb{S}$ as input and outputs the hidden states of each token:
\begin{equation}
\mathbf{Transformer}(\{s_{0},...,s_{L}\}) = \{h_{0},...,h_{L}\},
\end{equation}
where $h_{k}$ is the dense hidden state of $s_{k}$.

\noindent \textbf{Pre-training of PLMs}. This paper uses the auto-regression task as an instance of pre-training, which is commonly applied to auto-regressive PLMs \cite{radford2019language}. Given a sequence $\mathbb{S} = \{s_0, ... , s_L\}$, the goal is to model the joint probability of the sequence $ P(\mathbb{S})$.
\begin{equation}
     P(\mathbb{S}) = \prod_{k=1}^{L}{p(s_i|s_0,...s_{k-1})} 
\end{equation}
The transformer block is used to model these conditional probabilities. More specifically,
at step $k$ ($0<k\leq L$), the transformer receives $\{s_0...s_{k-1}\}$ and outputs their hidden states $\{h_{0}, ...,h_{k}\}$. The $h_{k}$ are used to predict the probability distribution of the next token.
\begin{equation}
p(s_i|s_0,...s_{k-1}) = \hat{s}_k = \sigma(\mathbf{Head}(h_{k}))
\end{equation}
The model parameters are trained to maximize the likelihood of $p(\mathbb{S})$, which is equivalent to minimizing the negative log-likelihood. Therefore, the loss function is:
\begin{equation}
    \mathcal{L}_{LM} = \sum_{k=1}^{L}{\textbf{CrossEntropy}(s_k,\hat{s}_k)}
\end{equation}

\noindent \textbf{Sentence representation}. 
Given a sentence $\mathbb{S}$ with length $L$, its sentence representation $W$ can be obtained by three methods \cite{reimers2019sentence, gao2021simcse}: (1) first token representation, which uses the hidden state of the $[\mathbf{CLS}]$ token ($h_{i,0}$) as sentence representation. 
(2) mean-pooling representation, which is obtained by mean-pooling of all hidden states (i.e., $\mathbf{Pool}(\{h_{0}...h_{L}\})$). (3) last token representation, which uses the hidden state of the last token. 

\noindent \textbf{PLMs with prompts}. 
Due to the gap between pretraining tasks and downstream tasks, sentence representation may be hard to contain all the sentence information, thereby requiring fine-tuning for specific tasks.
To address this issue, some studies utilize prompts to extract task-specific sentence features \cite{jiang2022promptbert}. 
For example, suppose a $\mathbb{S}_i$ is a paper titled ``Llama 2: Open Foundation and Fine-Tuned Chat Models'', and the task is to classify the subject of it belongs.
We can add some prompts to the sentence:
\begin{equation}
    \{[Title], this, paper, belong, to, which, subject?\}
\end{equation} 
We denote this new sentence with the prompt inserted as $\mathbb{S}_{i|\mathbb{P}}$, where $\mathbb{P}$ represents the newly inserted tokens. We use the hidden state of the last token as the sentence representation, denoted as $W_{i|\mathbb{P}}$. Since the last token is used to predict the next token distribution in the pre-training stage, it can naturally combine the inserted prompt information into the original sentence and extract the prompt-related semantics. Extensive studies \cite{liu2021p, lester2021power} show that using prompts can reduce the gap between PLMs and downstream tasks and maximize the utilization of knowledge learned by PLMs during pre-training.
\subsection{Graph Neural Network}

Graph Neural Networks (GNNs) have achieved remarkable success in modeling graphs \cite{velivckovic2017graph,gasteiger2018predict}. The message-passing framework is a commonly used architecture of GNN.

\noindent \textbf{Graph.} Let $G = \{V, A\}$ denote a graph, where $V$ is the node set and $A$ is the adjacency matrix, with $A_{ij}=1$ meaning there is an edge between node $i$ and node $j$. Usually, each node $i$ is associated with a node feature $x_i^{0}$.

\noindent \textbf{Framework of GNN.} The message-passing framework takes a set of node features $\mathcal{X} = \{x_i^0|i\in V\}$, and an adjacency matrix $A$ as input and iteratively captures neighbors' information via pooling. More specifically, for a given node $i \in V$ in the $l$-th layer of message-passing, it can be formulated as:
\begin{equation}
x_i^{l} = f_2(\mathbf{Pool}\{f_1(x^{l-1}_j|\theta_1^l) | j\in \mathcal{N}_i\},x_i | \theta_2^l)
\end{equation}
where $\mathbf{Pool}\{\cdot\}$ is an aggregation function that combines the features of neighboring nodes, such as mean-pooling. And $\mathcal{N}_i$ denotes the set of neighbors of node $i$. Besides, $f_1(\cdot|\theta_1^l)$ and $f_2(\cdot|\theta_2^l)$ denote two trainable transformations with parameters $\theta_1^l$ and $\theta_2^l$ respectively. Further, we denote an $l_{max}$ layer message-passing framework as GNN, formally:
\begin{equation}
z_i = \mathbf{GNN}(x_i^{0} , \mathcal{X}^0, A | \Theta_g)
\end{equation}
where $z_i = x_i^{l_{max}}$, and $\Theta_g$ represents all the trainable parameters in the GNN. We use $z_i$ as the structural representation for node $i$.

\subsection{Text-Attributed Graph}

Let $\mathcal{G} = \{\mathcal{V},\mathcal{A}\}$ denote a text-attributed graph, where $\mathcal{V}$ is the node set and $\mathcal{A}$ is the adjacency matrix. Each node $i \in \mathcal{V}$ is associated with a tokenized textual data, represented by $\mathbb{S}_i = \{s_{i,0},...,s_{i,L_i}\}$, which represents the textual data of the node. 
\\ 
\noindent \textbf{Problem Definition}: Given a text-attributed graph $\mathcal{G}$ and corresponding node labels $\mathcal{Y} = {y_i|i\in\mathcal{N}}$, this paper addresses the problem of efficiently modeling both the textual data ${\mathbb{S}_i | i\in\mathcal{V}}$ and the structural data in $\mathcal{G}$ to predict the node labels $\mathcal{Y}$.

\section{method}
This section introduces the proposed framework, referred to as \modelname, which uses GNNs as adapters for LLMs to better model TAGs.

\subsection{Overview}
\noindent \textbf{Motivation:} In the textual data of TAGs, many structure-related semantics are hard to infer from context alone. 
As illustrated in the example in Figure 1, we can easily infer that this user is ``popular'' based on his degree in the social network, but it is difficult to infer from their description of habits alone. 
Combining structural information can enhance language models' ability to model these structure-related semantics in TAGs. 
Meanwhile, the process of enhancement is learning how to model structure. 
Therefore, the proposed method \modelname, which first uses GNN as adapters for frozen PLMs, to combine structural information with PLMs, and then pre-trains them through the semantic understanding task on TAGs.
\\ \\
\noindent \textbf{Language-structure pre-training}: 
In the field of natural language processing, pre-training is a common strategy used to self-supervised enhance language models' ability for semantic understanding, with techniques such as auto-regressive pre-training (e.g., GPT-2/3 \cite{radford2019language, brown2020language}, Llama 2 \cite{touvron2023llama}, etc.) and auto-encoding pre-training (e.g., BERT\cite{yang2021bert}, RoBERTa\cite{liu2019roberta}, etc.).
Following our motivation, \modelname uses the same pre-training task as these PLMs. 
To facilitate comprehension, this section only discusses \modelname based on auto-regressive pre-training, and further details on how \modelname is combined with other pre-training tasks can be found in the appendix. 
Since the pre-training process uses the context semantic to supervise structure learning, we refer to this pre-training as language-structure pre-training.

\begin{figure*}[ht]
	\centering
	\includegraphics[width=0.95\textwidth]{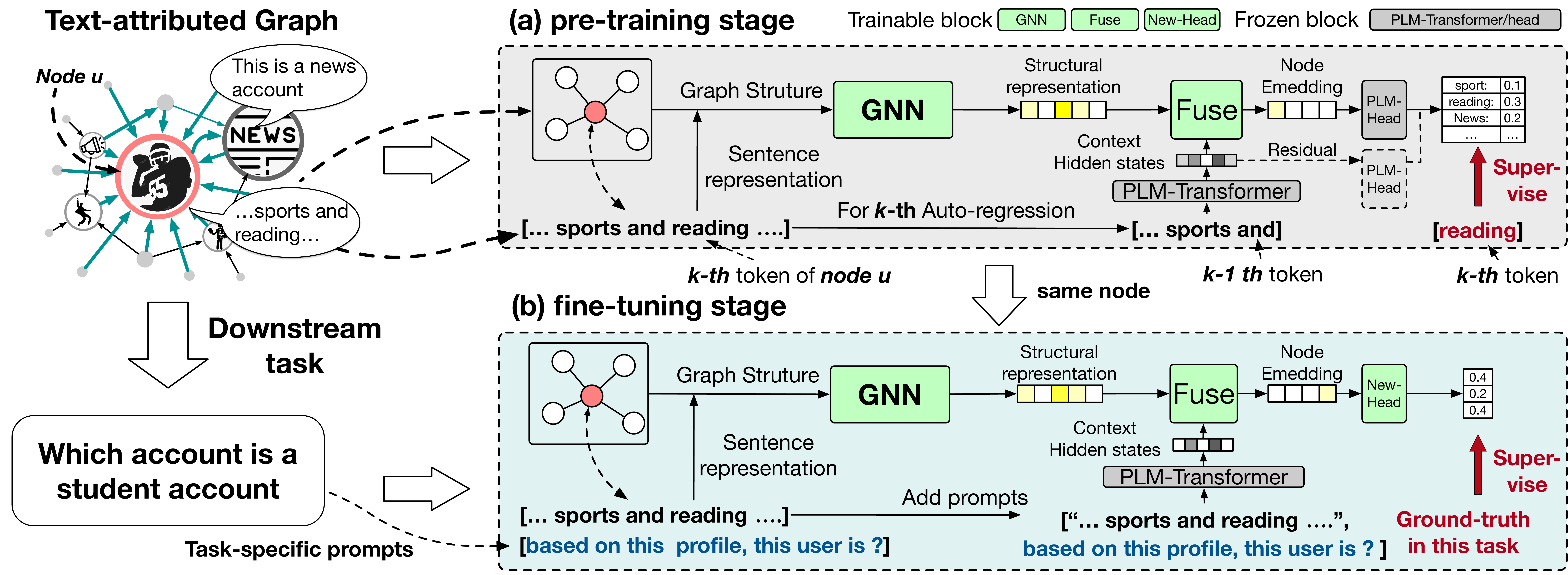}
	\caption{Framework of \modelname.
		\small In the pre-training stage, Step 1. GNN models the node structure information, Step 2. integrates the structural information with the corresponding text fragment encoded by LM, and Step 3. predicts the masked token. 
		}
	\label{fig:model:overview}
\end{figure*}

\noindent \\ \textbf{Framework:} The framework of \modelname is shown in Figure 2 (a). We also show how to fine-tune \modelname on the downstream tasks in Figure 2 (b), we detail this part in Section 4.3.
Given the textual data and graph structural data of a node, during the pre-training process, 
Step 1. GNN models the node structure information; Step 2. integrates the structural information with the corresponding context hidden-states modeled by PLM; and Step 3. predicts the next token. During this pre-training process, \modelname can learn rich information. \textbf{Align GNN with the language model.} During the learning process, the node representation obtained by GNN is constantly combined with different representations modeled by the language model for reasoning, and the entire process naturally aligns these two. \textbf{Enhance GNN in modeling graph structure.} During the entire pre-training stage, the semantic information in the textual data supervises the GNN to model the graph structural information.
\textbf{Better understanding the semantics in TAG}. GraphAapter can learn how to combine LLM and GNN to model the semantic information on TAG.

\subsection{Pre-training on TAGs}
In the training stage, \modelname uses the textual data of each node in TAG to train GNN.
\\ \\
\noindent \textbf{Pipeline of pre-training}: Given a text-attributed graph $\mathcal{G}$, node $i$ and its textual data $\mathbb{S}_i = \{s_{i,0},...,s_{i,L_i}\}$, \modelname uses all the tokens in $\mathbb{S}_i$ as supervision. 
For the $k$-th token, \modelname first extracts its previous tokens $\mathcal{S}_{i,k} = \{s_{i,0},...,s_{i,k-1}\}$. 
Then, GNN models node $i$'s structure information $z_i$. 
The structure information is then combined with the previous tokens to predict the probability distribution of the next token, where the ground truth is token $s_{i,k}$. 
\\ \\
\noindent \textbf{Structural representation}:
\modelname obtains its structural features $z_i$ through GNN. 
Here we use a general GNN based on the message-passing framework, which continuously aggregates neighbor features to obtain the new node's structural information. 
For whole process is formalized as:
\begin{equation}
z_i =  \mathbf{GNN}(x_i^{0}, \mathcal{X},\mathcal{A}| {\Theta_g})
\end{equation}
where $x_i^0$ and $\mathcal{A}$ represent the initial node feature input and adjacency matrix in GNN, respectively. This paper used the sentence representation of the corresponding node as $x_i^0$. See more details about GNN in Section 3.2.
\\ \\ 
\noindent \textbf{Context hidden-states}. 
\modelname use the pre-trained transformer in PLM to encode $\mathcal{S}_{i,k}$, it is formalized as:
\begin{equation}
    \label{model:input}
    h_{i,k} =  \textbf{Transformer}(\{s_{i,0}, s_{i,1},...,s_{i,k-1}\})
\end{equation}
Where the $\textbf{Transformer}$'s parameters are trained in frozen, and $h_{i,k}$ is the context hidden-states $\mathcal{S}_{i,k}$. 
Note that in the pretraining stage of PLM, $h_{i,k}$ is directly used to predict the next token, so $h_{i,k}$ contains both the context information and a certain of PLMs' prediction result.
\\ \\
\noindent \textbf{Fusion block}: 
\modelname next fuse structural representation into context hidden-states, which is formalized as:
\begin{equation}
    r_{i,k} =  \textbf{Fusion}(h_{i,k}, z_i | \Theta_{fuse}),
\end{equation}
The $\textbf{Fusion}(*)$ function is trainable with parameters $\Theta_{fuse}$. In this paper, MLPs are used as the structure of fusion. 
The process involves concatenating $h_{i,k}$ and $z_i$, and then feeding the resulting vector into MLPs.
\\ \\
\noindent \textbf{Residual connection}: the fused $r_{i,k}$ contains both structure information and context information. 
However, not every token's prediction requires the graph structure. 
For example, in the sentence "This paper focuses on graphs," the word "on" is simply a fixed collocation and easily inferred by context. 
Intuitively, words related to graph structure should be difficult for the language model to predict based on context. Therefore, the results of pre-trained language models are reused. 
We separately calculated the prediction probabilities of the language model alone and the probabilities that mixed the graph structure and the previous predictions. 
The two probabilities are then averaged to obtain the final prediction result.  
Formally:
\begin{equation}
    \hat{s}_{i, k}^{LM} = \sigma(\textbf{Head}(h_{i,k})), \hat{s}_{i, k}^{GNN} = \sigma(\textbf{Head}(r_{i,k}))
\end{equation}
\begin{equation}
    \label{equationResLabel}
    \hat{s}_{i, k}^{ALL} = (\hat{s}_{i, k}^{LM}+\hat{s}_{i, k}^{GNN})/2
\end{equation} 
Where $\sigma$ denotes the softmax function. 
Naturally, if a token $s_{i,k}$ can be accurately predicted by the language model and the corresponding score $\hat{s}_{i, k}^{GNN}$ is evenly distributed, the overall result remains correct. Conversely, if the token cannot be predicted by the language model, the \modelname~ needs to predict the correct token precisely to ensure the final result is correct. This difference leads the model to naturally focus on tokens originally predicted poorly by the language model during optimization. It then attempts to use additional structural data to enhance the overall framework's predictive performance.
\\ \\
\noindent \textbf{Optimization}: Our goal is to minimize the cross-entropy loss between the predicted probability distribution and the ground-truth distribution. 
Formally, 
\begin{equation}
    \mathcal{L}_{i,k} = \textbf{CrossEntropy}(\hat{s}_{i, k}^{ALL},s_{i,k})
\end{equation}

\begin{equation}
    \min_{\Theta_g,\Theta_{fuse}}{\sum_{i\in V}{\sum_{k \in \mathcal{S}_i}}{\mathcal{L}_{i,k}}}
\end{equation}
Note, only $\mathbf{GNN}(*|\Theta_G)$ and $\mathbf{Fusion}(*|\Theta_{fuse})$ of \modelname are trainable in whole pre-training.
\\ \\
\noindent \textbf{GNN as Adapter}: In the whole pre-training stage, the GNN 
 combines with the frozen LM's hidden states outputted from the transformer block. The combined hidden states are then input into the PLM's prediction head. Thus, the GNN acts as an adapter, altering the language model's predictions. Since the hidden states outputted by the transformer block can be pre-processed and stored in advance. Therefore, the entire training process only requires training the GNN. Therefore, GraphAdpater can efficiently pre-train based on different scales of PLMs.
\begin{table*}[t!]
    \caption{The performance of different methods across three datasets. 
        \small 
        Each row corresponds to a specific method, and each column presents the performance of the models on a particular dataset. The evaluation metric used is accuracy for the Arxiv and Reddit datasets, and ROC-AUC for Instagram. The LM employed in each method is indicated in parentheses.
        \normalsize 
        } 
 	\normalsize
    \begin{tabular}{c|c|c|c|c}
    \toprule
     & & {\textbf{Arxiv}} & {\textbf{Instagram}} & {\textbf{Reddit}} \\
\midrule
 \multirow{7}{*}{LM}& GNN (Ogb-feature) &  0.6980 \tiny{(0.0013)}&  -                  & - \\
& GNN (RoBERTa)         &  0.7129 \tiny{(0.0013)}  &  0.6123 \tiny{(0.0063)}       & 0.6191 \tiny{(0.0043)} \\

& GNN (RoBERTa+Prop)           & 0.7067 \tiny{(0.0011)} &  0.6138 \tiny{(0.0117)} & 0.6198 \tiny{(0.0036)} \\

& GIANT (BERT)                 & 0.7262 \tiny{(0.0011)} & 0.5986 \tiny{(0.0022)} & 0.6379 \tiny{(0.0045)}\\

& GIANT (BERT+Prop)            & 0.7252 \tiny{(0.0012)} & 0.6029 \tiny{(0.0123)} & 0.6348 \tiny{(0.0039)}\\
& GLEM$^1$ (DeBERTa)           & 0.7550 \tiny{(0.0024)}               & - & - \\
& GLEM (DeBERTa)               & 0.7355 \tiny{(0.0034)}                  & 0.6166 \tiny{(0.0056)} & 0.6228 \tiny{(0.0060)} \\

& GLEM (DeBERTa+Prop)          & 0.7315 \tiny{(0.0033)}                 & 0.6105 \tiny{(0.0038)} & 0.6221 \tiny{(0.0052)} \\
\hline
\multirow{7}{*}{LLM}&MLPs (Llama 2 + Prop) & 0.7541 \tiny{(0.0024)} & 0.6248 \tiny{(0.0111)} & 0.6123 \tiny{(0.0034)}\\
 &LoRA (Llama 2 + Prop) &0.7454 \tiny{(0.0012)} &0.5910 \tiny{(0.0160)} & 0.5740 \tiny{(0.0172)}\\
 &  GNN (Llama 2)
                               & 0.7305 \tiny{(0.0020)} & 0.6221 \tiny{(0.0112)} & 0.6320 \tiny{(0.0041)} \\

& GNN (Llama 2+Prop)           & 0.7336 \tiny{(0.0018)} & 0.6312 \tiny{(0.0051)} & 0.6324 \tiny{(0.0033)}\\
& Graph2Text (Llama 2 + Prop) & 0.7348 \tiny{(0.0026)}& 0.6226 \tiny{(0.0045)}&0.6053 \tiny{(0.0033)} \\
& TAPE (GPT-3.5)                & 0.7672 \tiny{(0.0007)}
                                                        & -                      & -\\ \hline

\multirow{2}{*}{Ours}& GraphAdapter (w/o Pre)                 & 0.7648 \tiny{(0.0020)} & 0.6351 \tiny{(0.0077)} & 0.6369 \tiny{(0.0025)}
	\\ 
& \textbf{GraphAdapter}                 & \textbf{0.7707 \tiny{(0.0015)}} & \textbf{0.6513 \tiny{(0.0075)}} & \textbf{0.6461 \tiny{(0.0019)}}
	\\ 
 \bottomrule
    \end{tabular}
\label{tbl:main}

\footnotemark{performance reported in \cite{zhao2022learning}}
\end{table*} 
\subsection{Fine-tuning with Prompts}
The pipeline is shown in Figure 2 (b).
\modelname is pre-trained by token-level semantic understanding tasks. 
To better utilize the learned knowledge of \modelname and the PLMs in downstream tasks, we further proposed prompt-aware fine-tuning. 
It inserts prompts in textual data to get task-specific sentence embedding of each node. 
Prompts can transform various downstream tasks on TAGs into next token prediction. 
E.g., the task ``\textit{Which account is a
student account}'' can be transformed by a next-token prediction task, ``[context], \textit{based on this profile, this user is }''. 
In the pre-training stage, \modelname has learned how to utilize the structural information captured by GNN to enhance the accuracy of next-token prediction, therefore, under the transformed downstream task can better utilize the learned knowledge from pre-training.
Formally, given textual data $\mathbb{S}_i$ of node $i$, we can combine a sequence of tokens with task-specific prompts behind textual data, namely, $\mathbb{S}_{i|\mathbb{P}}= [\mathbb{S}_i,\mathbb{P}]$, then we can get its sentence hidden states $h_{i|\mathbb{P}}$ through the transformer of PLM.
The resulting hidden state is then fused with the node's structural representation as node representation in a specific downstream task.
\begin{equation}
    r_{i|\mathbb{P}} = \mathbf{Fusion}(h_{i|\mathbb{P}},z_i)
\end{equation}
This node representation can be used in various tasks. For example, in the node classification, we can append a new linear transformation to output the result, i.e., $\hat{y}_{i|\mathbb{P}}=f(r_{i|\mathbb{P}}|\theta_{new})$. In fine-tuning stage, the whole parameters $\{\Theta_g,\Theta_{fuse}, \theta_{new}\}$ in \modelname are trainable.

\section{Experiment}

\begin{table*}[ht!]
    \caption{The performance of the GraphAdapter based on different LM across three datasets. 
    \small 
    The evaluation metrics used for these datasets align with those outlined in Table \ref{tbl:main}.
    \normalsize
    }
    \normalsize
\resizebox{0.95\textwidth}{!}{
\begin{tabular}{c|ccc|ccc|ccc}
    \toprule
 & \multicolumn{3}{c|}{\textbf{Arxiv}} & \multicolumn{3}{c|}{\textbf{Instagram}} & \multicolumn{3}{c}{\textbf{Reddit}} \\
 &RoBERTa & GPT2 &Llama 2 &RoBERTa & GPT-2 &Llama 2&RoBERTa & GPT-2 &Llama 2 \\
\midrule

GNN (PLM)                   & 0.7129 \tiny{(0.0013)} & 0.7174 \tiny{(0.0019)} & 0.7305 \tiny{(0.0022)}
                             & 0.6123 \tiny{(0.0063)} & 0.6019 \tiny{(0.0124)} & 0.6221 \tiny{(0.0112)} 
                             & 0.6191\tiny{(0.0043)}  & 0.6282 \tiny{(0.0036)} & 0.6320 \tiny{(0.0041)}  \\
\hline

GNN (PLM+Prop)                       &0.7067 \tiny{(0.0011)} &0.6915 \tiny{(0.0021)} &  0.7336 \tiny{(0.0027)} 
                             & 0.6138 \tiny{(0.0117)} & 0.6128 \tiny{(0.0014)} & 0.6312 \tiny{(0.0051)}  
                             & 0.6198 \tiny{(0.0036)} & 0.6206 \tiny{(0.0011)} & 0.6324 \tiny{(0.0033)}  \\
\hline

GraphAdapter (w/o Pre) &  0.7069 \tiny{(0.0026)}  & 0.7146 \tiny{(0.0025)} & 0.7648 \tiny{(0.0020)}
                             & 0.6165 \tiny{(0.0038)} & 0.6162 \tiny{(0.0066)} & 0.6351 \tiny{(0.0077)}  
                             & 0.6210 \tiny{(0.0036)} & 0.6284 \tiny{(0.0027)} & 0.6369 \tiny{(0.0025)}  \\
\hline

GraphAdapter                 & \textbf{0.7273 \tiny{(0.0021)}}  & \textbf{0.7325 \tiny{(0.0022)}} & \textbf{0.7707 \tiny{(0.0015)}}
                             & \textbf{0.6292 \tiny{(0.0033)}} & \textbf{0.6276 \tiny{(0.0034)}} & \textbf{0.6508 \tiny{(0.0033)}}
                             & \textbf{0.6379 \tiny{(0.0061)}} & \textbf{0.6441 \tiny{(0.0022)}} & \textbf{0.6461 \tiny{(0.0019)} }
	\\ \bottomrule
    \end{tabular}
    }
\label{tbl:difflm}
\end{table*} 

To comprehensively validate that \modelname can mine the intrinsic correlation between the textual and structure data in TAGs, we conduct extensive experiments on three real-world datasets from diverse domains.

Our experimentation centered on the following five questions:
\begin{itemize}[leftmargin=*]
    \item \textbf{\textit{Q1}: How well is \modelname in modeling TAGs?}
    \item \textbf{\textit{Q2}: Whether \modelname can adapt to other PLMs?}
    \item \textbf{\textit{Q3}: Are all components comprising \modelname valid?}
    \item \textbf{\textit{Q4}: What exactly does \modelname's pre-training learn?}
    \item \textbf{\textit{Q5}: How efficient is GraphAdapter?}
\end{itemize}

\subsection{Experiment setup}

\textbf{Dataset and metrics.} We select three public and real-world datasets used for evaluation: Ogbn-arxiv \cite{hu2020open} (shorted as Arxiv), Instagram \cite{kim2020multimodal} and Reddit\footnote{\url{https://convokit.cornell.edu/documentation/subreddit.html}}. The evaluation task involves node classification. The metric used for Arxiv and Reddit is accuracy, while for Instagram, it is ROC-AUC. See more details in Appendix \ref{appendix:data}.
\\
\noindent \textbf{Baselines.} We compare the proposed \modelname with several TAG modeling methods. They are LM+MLPs, LM+GNN, GIANT \cite{chien2021node}, GLEM \cite{zhao2022learning}, Graph2Text \cite{chen2023exploring}, LoRA \cite{hu2021lora} and TAPE \cite{he2023explanations}. Since most of these methods can combine with different GNN blocks and PLMs, and the specific GNN framework is not the key point this paper focuses on, this paper uses GraphSAGE \cite{hamilton2017inductive} as an instance of GNN. And we detail the used PLMs in Table \ref{tbl:main}. Please refer to Appendix \ref{appendix:baselines} for more details. \\ 
\noindent \textbf{Prompts.} Since \modelname involves prompts, to make a fair comparison, we also enhance the baselines with prompts (denoted as ``+Prop''). For further details, please refer to the Appendix \ref{appendix:prompt}. 
\\ 
\noindent \textbf{Implementation details.} In the experiment, Llama 2 defaulted to the 13B version, while other language models defaulted to the large version. For further details, please refer to the Appendix \ref{appendix:implementation}.

\subsection{Performance}
\textbf{\textit{Q1}: How well is \modelname in modeling TAGs?}

\textbf{\textit{A1}: \modelname can effectively model TAGs and surpass current state-of-the-art baselines on node classification tasks.} We compare \modelname with 6 state-of-the-art baselines on 3 different real-world datasets to evaluate its effectiveness. As Table \ref{tbl:main} shows, the experiment results suggest:
\\
(1) Frozen LLMs are effective on TAGs. 
In general, frozen LLMs have an improved performance compared to the previous frozen LM. 
Experiment results show Llama 2 has improved performance on 3 datasets by 1.34\% compared to RoBERTa-based methods. LLM can better combine the information in prompts to extract task-relevant sentence representations of nodes. As the results show, prompts can bring a 0.42\% improvement on average for LLM, but they could not improve the performance of LM. Frozen LLMs with prompt can surpass many GNN-LM methods that require tuning LM. Results also show that LLMs with prompts can surpass GLEM and GIANT by 0.43\% and 0.79\% on average, respectively.
\\
(2) Directly fusing GNN and LLM results in unstable improvements. Compared to ordinary GNN, \modelname (w/o Pre) only adds one fusion component to fuse the semantic representation from the LM and structural representation from the GNN. Experiment results show that directly fusing language model representations only brings improvements on Arxiv, but not obviously on other datasets. Note that the Arxiv training samples are much larger than the other datasets. This result suggests that training samples may have an impact on GNNs to understand and effectively incorporate the representations inferred by LLMs with prompts.
\\
(3) \modelname can effectively combine GNN and LLM, surpassing existing state-of-the-art baselines in terms of performance. The pre-training effect of \modelname is significant, bringing an average performance improvement of 1.98\% and thus surpassing existing state-of-the-art baselines. Specifically, \modelname achieves an improvement of 4.72\% over state-of-the-art cascaded GNN-LM methods and 5.40\% over self-supervised GNN-LMs on average. At the same time, \modelname also surpasses TAPE, another LLM-based method on Arxiv by 0.4\% accuracy improvement.
\\ \\
\noindent\textbf{\textit{Q2}: Whether \modelname can adapt to other PLMs?} 

\textbf{\textit{A2}: \modelname can be effectively pre-trained based on RoB-ERTa, GPT-2, and Llama 2, resulting in performance improvements.} We run GraphAapter based on 3 different LMs. The experiment results are shown in Table \ref{tbl:difflm}. \modelname improved average performance over directly combining GNNs with frozen PLM by 1.67\% on RoBERTa, 1.89\% on GPT-2, and 2.77\% on Llama 2. Meanwhile, \modelname pre-training brings 1.67\%, 1.50\%, and 1.02\% improvements on RoBERTa, GPT-2, and Llama 2 respectively. This result fully demonstrates that \textbf{\modelname is a general and scalable method.} It is worth noting that the pre-training method of RoBERTa is different from others. \modelname uses a pre-training task similar to RoBERTa, so there are some slight differences from the formula in Section 4. The main differences come from the loss function and language model inputs. We describe the details of applying \modelname on Roberta in the appendix.

\textbf{Under the same PLM, the performance of \modelname is comparable to the SOTA baselines based on fine-tuning the PLM.} We evaluate the performance between \modelname and SOTA baselines under the same LM. Since the GLEM adopted DeBERTa, however, the pre-training code of DeBERTa is not open-sourced at present. To keep consistent, \modelname and GLEM both adopt the same RoBERTa-base. As shown in Table \ref{tbl:same_lm}, the experiment results suggest that methods based on pre-training like GIANT and \modelname perform better on small datasets like Instagram and Reddit. Similarly, Roberta-based \modelname outperforms GLEM by 1.57\% and BERT-based GIANT outperforms GLEM by 1.15\% on small datasets. 
Compared to baselines based on pre-training, although GIANT fine-tunes the LM, its performance is 0.51\% lower than \modelname on average.
Therefore, overall, even without fine-tuning the LM, the performance of \modelname is comparable to current state-of-the-art baselines based on fine-tuning the LM.
\begin{table}[]
    \caption{The performance of different methods using the same LMs across three datasets. 
    \small
    The evaluation metrics employed for these datasets align with those described in Table \ref{tbl:main}.
    \normalsize
    }
 	\normalsize
  
\resizebox{0.45\textwidth}{!}{
    \begin{tabular}{c|c|c|c}
    \toprule
      & {\textbf{Arxiv}} & {\textbf{Instagram}} & {\textbf{Reddit}} \\
\midrule
GNN (BERT)            & 0.7039  \tiny{(0.0013)} & 0.5973 \tiny{(0.0063)} & 0.6061 \tiny{(0.0043)} \\
 \hline
GIANT (BERT)            & \textbf{0.7269 \tiny{(0.0021)}} & 0.5986 \tiny{(0.0022)} & \textbf{0.6379 \tiny{(0.0045)}} \\
 \hline
GraphAdapter (BERT)   & 0.7264 \tiny{(0.0012)} & \textbf{0.6156  \tiny{(0.0032)}} & 0.6366 \tiny{(0.0034)} \\

 \hline
 \hline
 GNN (RoBERTa)        & 0.7129 \tiny{(0.0013)} & 0.6123  \tiny{(0.0063)} & 0.6191 \tiny{(0.0043)} \\
 \hline
 GLEM (RoBERTa)         & \textbf{0.7308 \tiny{(0.0029)}}& 0.6114 \tiny{(0.0075)} & 0.6228 \tiny{(0.0018)} \\
 \hline
GraphAdapter (RoBERTa)  & 0.7273 \tiny{(0.0021)} & \textbf{0.6276  \tiny{(0.0034)}} & \textbf{0.6379 \tiny{(0.0061)}}	\\ \bottomrule
    \end{tabular}}
\label{tbl:same_lm}
\end{table} 

\subsection{In-depth Analysis.}
\textbf{\textit{Q3}: Are all components comprising \modelname valid?}

\textbf{\textit{A3}: As Table \ref{tbl:ablation} shows, removing any component of Grap-hAdapter results in performance drops. }
Removing pre-training leads to a 0.91\% drop, demonstrating that \modelname's improvements indeed come from pre-training. 
Next, the most significant performance drop is when we simultaneously remove pre-training and graph structure in the fine-tuning stage (keeping only self-loops), which causes a 1.95\% drop. This shows having the graph is crucial for \modelname to work. 
Removing the task-related prompt leads to a 0.98\% drop, validating our design of aligning pre-training tasks via prompts. 
Notably, removing the residual learning (``w/o Res Label'' that is stated in Equation \ref{equationResLabel})  leads to a 1.02\% drop (more than removing pre-training), suggesting that training GNNs directly on all text may introduce excessive noise and hurt performance. 
Therefore, \modelname indeed benefited from residual learning, which utilizes language model predictions to select words more semantically related to the graph. \\ \\
\noindent\textbf{\textit{Q4}: What exactly does \modelname's pre-training learn?}
\begin{figure}
	\centering
	\includegraphics[width=0.48\textwidth]{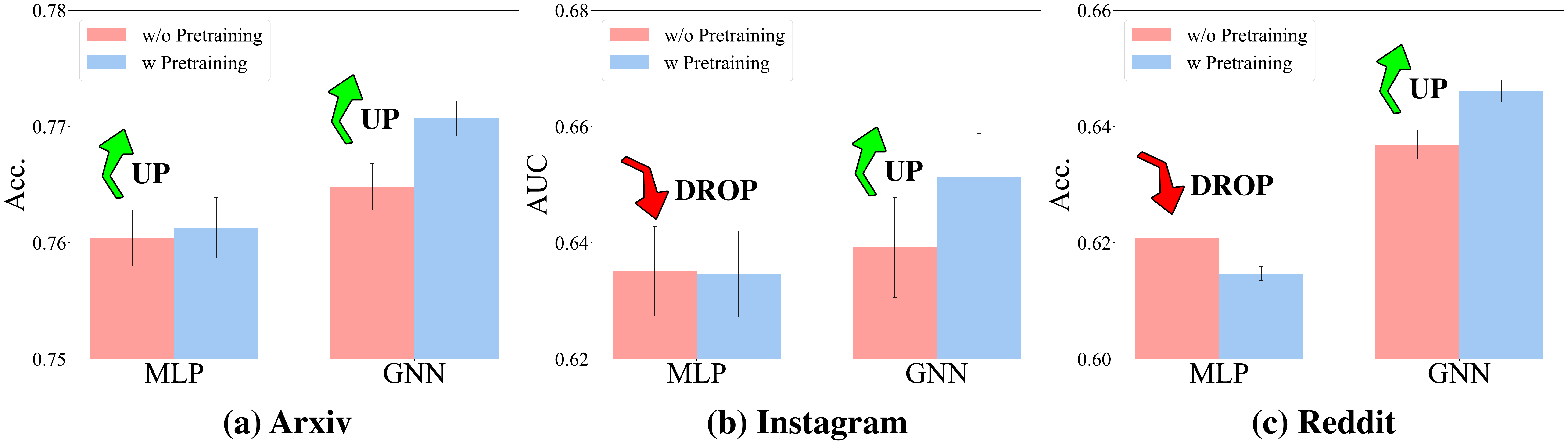}
	\caption{The performance of \modelname before and after pre-training, using MLP and GNN as the backbone architectures. 
        \small
        The red represents performance without pre-training, while the blue represents performance after pre-training.
        \normalsize
        }
        \normalsize
	\label{fig:diffblck}
\end{figure}  
We conduct three comparative experiments to demonstrate what \modelname pre-training is doing.

\textbf{(1) GNN can obtain stronger expressive power through pre-training.} 
We first observe the performance change of GNNs before and after pre-training, where we directly use the structural representations from the pre-trained GNN to fine-tune for downstream tasks. 
As Table \ref{tbl:gnn} shows, the pre-trained GNN performs better on downstream tasks, improving by 0.78\% on average. 
This demonstrates that GNNs are training their ability to model the graph structure during pre-training.

\textbf{(2) Fusion block is learning how to fuse the knowledge from the language model and GNN during pre-training.}  We further explore whether the fusion layer learned useful knowledge during training. We randomly initialize the parameters in a specific \modelname's blocks after pre-trained. As Table \ref{tbl:initi} shows, initializing the parameters of the fusion layer leads to significant performance drops, decreasing by 1.03\% on average across 3 datasets. This result shows that the enhanced knowledge from GNN may need to be outputted through the matching fusion layer. To further verify this conjecture, we further reinitialized the parameters of GNN, and some performance decline can also be observed, decreasing by 0.82\% on average. This is similar to the impact of reinitializing the fusion layer. The fusion layer alone does not contain much knowledge. Therefore, these results demonstrate that the fusion layer can learn how to fuse the knowledge from GNN and language models.

\textbf{(3) Graph structure is the basis of pre-training.} We further observe the changes in different base models before and after pre-training. In this comparative experiment, we keep all the structures of \modelname, only replacing the GNN block with MLPs of equal parameter size. As Figure \ref{fig:diffblck} shows, the MLP-based \modelname shows no significant change before and after pre-training (average improvement of 0.19\%), and even decreases in performance on Instagram and Reddit (drops of 0.05\% and 0.62\% respectively). While the GNN improves notably before and after pre-training (average improvement of 0.91\%). This result suggests that GNN is a prerequisite for effective pre-training.

These three results demonstrate that \modelname is indeed learning graph structures via pre-training. This validates that the pre-training of \modelname is reasonable and effective, and further supports the motivation of this paper. 
\\ \\ 
Furthermore, we investigate the performance of \modelname with different GNN blocks (see Appendix \ref{appendix:analysis_prormpts}) and conduct more detailed ablation studies (see Appendix \ref{appendix:ablation}). Additionally, we analyze and report the efficiency of \modelname (see Appendix \ref{appendix:efficent}) to answer the \textit{\textbf{Q5}}. Moreover, we conduct a case study on Arxiv to further demonstrate the advantages of the proposed method (see Appendix \ref{appendix:case}).

\begin{table}[]
    \caption{The performance of \modelname when various components are removed. 
    \small
    The evaluation metrics used for these tests align with those described above. The term 'w/o' indicates removing a specific component from the \modelname.
    \normalsize
    }
 	\normalsize
  
\resizebox{0.45\textwidth}{!}{
    \begin{tabular}{c|c|c|c}
    \toprule
      & {\textbf{Arxiv}} & {\textbf{Instagram}} & {\textbf{Reddit}} \\
\midrule
w/o Pretraining          & 0.7648 \tiny{(0.0020)} & 0.6392 \tiny{(0.0086)} & 0.6369 \tiny{(0.0025)}  \\
 \hline
w/o Graph structure      & 0.7604 \tiny{(0.0024)} & 0.6346 \tiny{(0.0074)} & 0.6147 \tiny{(0.0012)} \\
\hline
w/o Res label            & 0.7605 \tiny{(0.0013)}  & 0.6408 \tiny{(0.0130)} & 0.6363 \tiny{(0.0036)}  \\
 \hline
w/o task-specific prompt  & 0.7594 \tiny{(0.0030)}  & 0.6364 \tiny{(0.0073)} & 0.6430 \tiny{(0.0021)} \\
\hline
GraphAdapter             & 0.7707 \tiny{(0.0015)} & 0.6513 \tiny{(0.0075)} & 0.6461 \tiny{(0.0019)} 

	\\ \bottomrule
    \end{tabular}
\label{tbl:ablation}}
\end{table}

\begin{table}[]
    \caption{The performance changes of the GNN block in \modelname before and after pre-training. 
    \small
    Here, ``w/o Pretraining'' signifies no pre-training, while ``w Pretraining'' indicates the opposite.
    \normalsize
    }
    \normalsize
  
\resizebox{0.45\textwidth}{!}{
    \begin{tabular}{c|c|c|c}
    \toprule
      & {\textbf{Arxiv}} & {\textbf{Instagram}} & {\textbf{Reddit}} \\
\midrule
GNN w/o Pretraining     & 0.7305  \tiny{(0.0020)} & 0.6181 \tiny{(0.0112)} & 0.6320 \tiny{(0.0041)} \\
 \hline
GNN w Pretraining     & \textbf{0.7335 \tiny{(0.0024)}} & \textbf{0.6294 \tiny{(0.0038)}} & \textbf{0.6410 \tiny{(0.0027)}} \\
\bottomrule
    \end{tabular}
\label{tbl:gnn}}
\end{table}

\begin{table}[]
    \caption{The performance of \modelname after randomly initializing some blocks. 
    \small
    Here, ''Re-init'' represents re-initialization.
    \normalsize
    }
 	\normalsize
  
\resizebox{0.45\textwidth}{!}{
    \begin{tabular}{c|c|c|c}
    \toprule
      & {\textbf{Arxiv}} & {\textbf{Instagram}} & {\textbf{Reddit}} \\
\midrule
Re-init All          & 0.7648 \tiny{(0.0020)} & 0.6392 \tiny{(0.0086)} & 0.6369 \tiny{(0.0025)}  \\ \hline
Re-init GNN          & 0.7680 \tiny{(0.0022)} & 0.6390 \tiny{(0.0050)} & 0.6364 \tiny{(0.0026)}  \\ \hline
 Re-init Fusion   & 0.7562 \tiny{(0.0011)} & 0.6431 \tiny{(0.0024)} & 0.6378 \tiny{(0.0022)} \\
 \hline
GraphAdapter             & 0.7707 \tiny{(0.0015)} & 0.6513 \tiny{(0.0075)} & 0.6461 \tiny{(0.0019)}  \\
\bottomrule
    \end{tabular}
\label{tbl:initi}}
\end{table}

\section{CONCLUSION}
 This paper proposes \modelname~to harness LLMs for TAGs without fine-tuning. A GNN adapter is trained to reduce LLM next-word errors on node texts. This adapts LLMs for graphs efficiently. Across node classification tasks, \modelname improves accuracy by 5\% over baselines. We validate with RoBERTa, GPT-2, and Llama 2, efficiently leveraging LLMs for interconnected text-graph data.
\section*{ACKNOWLEDGMENTS}
This work is supported by Natural Science
Foundation of China (No.62322606).
\newpage
\balance
\bibliographystyle{ACM-Reference-Format}
\bibliography{main}


\begin{thebibliography}{43}

%%% the Web address.

\ifx \showCODEN    \undefined \def \showCODEN     #1{\unskip}     \fi
\ifx \showDOI      \undefined \def \showDOI       #1{#1}\fi
\ifx \showISBNx    \undefined \def \showISBNx     #1{\unskip}     \fi
\ifx \showISBNxiii \undefined \def \showISBNxiii  #1{\unskip}     \fi
\ifx \showISSN     \undefined \def \showISSN      #1{\unskip}     \fi
\ifx \showLCCN     \undefined \def \showLCCN      #1{\unskip}     \fi
\ifx \shownote     \undefined \def \shownote      #1{#1}          \fi
\ifx \showarticletitle \undefined \def \showarticletitle #1{#1}   \fi
\ifx \showURL      \undefined \def \showURL       {\relax}        \fi
% invisible to TeX
\providecommand\bibfield[2]{#2}
\providecommand\bibinfo[2]{#2}
\providecommand\natexlab[1]{#1}
\providecommand\showeprint[2][]{arXiv:#2}

\bibitem[Berge(2001)]%
        {berge2001theory}
\bibfield{author}{\bibinfo{person}{Claude Berge}.}
  \bibinfo{year}{2001}\natexlab{}.
\newblock \showarticletitle{The theory of graphs}. In
  \bibinfo{booktitle}{\emph{Courier Corporation}}.
\newblock


\bibitem[Brown et~al\mbox{.}(2020)]%
        {brown2020language}
\bibfield{author}{\bibinfo{person}{Tom Brown}, \bibinfo{person}{Benjamin Mann},
  \bibinfo{person}{Nick Ryder}, \bibinfo{person}{Melanie Subbiah},
  \bibinfo{person}{Jared~D Kaplan}, \bibinfo{person}{Prafulla Dhariwal},
  \bibinfo{person}{Arvind Neelakantan}, \bibinfo{person}{Pranav Shyam},
  \bibinfo{person}{Girish Sastry}, \bibinfo{person}{Amanda Askell},
  {et~al\mbox{.}}} \bibinfo{year}{2020}\natexlab{}.
\newblock \showarticletitle{Language models are few-shot learners}. In
  \bibinfo{booktitle}{\emph{NeurIPS'20}}, Vol.~\bibinfo{volume}{33}.
  \bibinfo{pages}{1877--1901}.
\newblock


\bibitem[Chen et~al\mbox{.}(2023)]%
        {chen2023exploring}
\bibfield{author}{\bibinfo{person}{Zhikai Chen}, \bibinfo{person}{Haitao Mao},
  \bibinfo{person}{Hang Li}, \bibinfo{person}{Wei Jin},
  \bibinfo{person}{Hongzhi Wen}, \bibinfo{person}{Xiaochi Wei},
  \bibinfo{person}{Shuaiqiang Wang}, \bibinfo{person}{Dawei Yin},
  \bibinfo{person}{Wenqi Fan}, \bibinfo{person}{Hui Liu}, {et~al\mbox{.}}}
  \bibinfo{year}{2023}\natexlab{}.
\newblock \showarticletitle{Exploring the potential of large language models
  (llms) in learning on graphs}. In \bibinfo{booktitle}{\emph{arXiv preprint
  arXiv:2307.03393}}.
\newblock


\bibitem[Chien et~al\mbox{.}(2021)]%
        {chien2021node}
\bibfield{author}{\bibinfo{person}{Eli Chien}, \bibinfo{person}{Wei-Cheng
  Chang}, \bibinfo{person}{Cho-Jui Hsieh}, \bibinfo{person}{Hsiang-Fu Yu},
  \bibinfo{person}{Jiong Zhang}, \bibinfo{person}{Olgica Milenkovic}, {and}
  \bibinfo{person}{Inderjit~S Dhillon}.} \bibinfo{year}{2021}\natexlab{}.
\newblock \showarticletitle{Node feature extraction by self-supervised
  multi-scale neighborhood prediction}. In \bibinfo{booktitle}{\emph{arXiv
  preprint arXiv:2111.00064}}.
\newblock


\bibitem[Corley et~al\mbox{.}(2010)]%
        {corley2010text}
\bibfield{author}{\bibinfo{person}{Courtney~D Corley}, \bibinfo{person}{Diane~J
  Cook}, \bibinfo{person}{Armin~R Mikler}, {and} \bibinfo{person}{Karan~P
  Singh}.} \bibinfo{year}{2010}\natexlab{}.
\newblock \showarticletitle{Text and structural data mining of influenza
  mentions in web and social media}. In \bibinfo{booktitle}{\emph{International
  journal of environmental research and public health}},
  Vol.~\bibinfo{volume}{7}. \bibinfo{pages}{596--615}.
\newblock


\bibitem[Devlin et~al\mbox{.}(2018)]%
        {devlin2018bert}
\bibfield{author}{\bibinfo{person}{Jacob Devlin}, \bibinfo{person}{Ming-Wei
  Chang}, \bibinfo{person}{Kenton Lee}, {and} \bibinfo{person}{Kristina
  Toutanova}.} \bibinfo{year}{2018}\natexlab{}.
\newblock \showarticletitle{Bert: pre-training of deep bidirectional
  transformers for language understanding}. In \bibinfo{booktitle}{\emph{arXiv
  preprint arXiv:1810.04805}}.
\newblock


\bibitem[Duan et~al\mbox{.}(2023)]%
        {duan2023simteg}
\bibfield{author}{\bibinfo{person}{Keyu Duan}, \bibinfo{person}{Qian Liu},
  \bibinfo{person}{Tat-Seng Chua}, \bibinfo{person}{Shuicheng Yan},
  \bibinfo{person}{Wei~Tsang Ooi}, \bibinfo{person}{Qizhe Xie}, {and}
  \bibinfo{person}{Junxian He}.} \bibinfo{year}{2023}\natexlab{}.
\newblock \showarticletitle{Simteg: A frustratingly simple approach improves
  textual graph learning}. In \bibinfo{booktitle}{\emph{arXiv preprint
  arXiv:2308.02565}}.
\newblock


\bibitem[Gao et~al\mbox{.}(2021)]%
        {gao2021simcse}
\bibfield{author}{\bibinfo{person}{Tianyu Gao}, \bibinfo{person}{Xingcheng
  Yao}, {and} \bibinfo{person}{Danqi Chen}.} \bibinfo{year}{2021}\natexlab{}.
\newblock \showarticletitle{Simcse: simple contrastive learning of sentence
  embeddings}. In \bibinfo{booktitle}{\emph{arXiv preprint arXiv:2104.08821}}.
\newblock


\bibitem[Gasteiger et~al\mbox{.}(2018)]%
        {gasteiger2018predict}
\bibfield{author}{\bibinfo{person}{Johannes Gasteiger},
  \bibinfo{person}{Aleksandar Bojchevski}, {and} \bibinfo{person}{Stephan
  G{\"u}nnemann}.} \bibinfo{year}{2018}\natexlab{}.
\newblock \showarticletitle{Predict then propagate: Graph neural networks meet
  personalized pagerank}. In \bibinfo{booktitle}{\emph{arXiv preprint
  arXiv:1810.05997}}.
\newblock


\bibitem[Guo et~al\mbox{.}(2023)]%
        {guo2023gpt4graph}
\bibfield{author}{\bibinfo{person}{Jiayan Guo}, \bibinfo{person}{Lun Du}, {and}
  \bibinfo{person}{Hengyu Liu}.} \bibinfo{year}{2023}\natexlab{}.
\newblock \showarticletitle{GPT4Graph: can large language models understand
  graph structured Data? An Empirical Evaluation and Benchmarking}. In
  \bibinfo{booktitle}{\emph{arXiv preprint arXiv:2305.15066}}.
\newblock


\bibitem[Hamilton et~al\mbox{.}(2017)]%
        {hamilton2017inductive}
\bibfield{author}{\bibinfo{person}{Will Hamilton}, \bibinfo{person}{Zhitao
  Ying}, {and} \bibinfo{person}{Jure Leskovec}.}
  \bibinfo{year}{2017}\natexlab{}.
\newblock \showarticletitle{Inductive representation learning on large graphs}.
  In \bibinfo{booktitle}{\emph{NeurIPS'17}}, Vol.~\bibinfo{volume}{30}.
\newblock


\bibitem[He et~al\mbox{.}(2020)]%
        {he2020deberta}
\bibfield{author}{\bibinfo{person}{Pengcheng He}, \bibinfo{person}{Xiaodong
  Liu}, \bibinfo{person}{Jianfeng Gao}, {and} \bibinfo{person}{Weizhu Chen}.}
  \bibinfo{year}{2020}\natexlab{}.
\newblock \showarticletitle{Deberta: decoding-enhanced bert with disentangled
  attention}. In \bibinfo{booktitle}{\emph{arXiv preprint arXiv:2006.03654}}.
\newblock


\bibitem[He et~al\mbox{.}(2023)]%
        {he2023explanations}
\bibfield{author}{\bibinfo{person}{Xiaoxin He}, \bibinfo{person}{Xavier
  Bresson}, \bibinfo{person}{Thomas Laurent}, {and} \bibinfo{person}{Bryan
  Hooi}.} \bibinfo{year}{2023}\natexlab{}.
\newblock \showarticletitle{Explanations as Features: LLM-Based Features for
  text-attributed graphs}. In \bibinfo{booktitle}{\emph{arXiv preprint
  arXiv:2305.19523}}.
\newblock


\bibitem[Hu et~al\mbox{.}(2021)]%
        {hu2021lora}
\bibfield{author}{\bibinfo{person}{Edward~J Hu}, \bibinfo{person}{Yelong Shen},
  \bibinfo{person}{Phillip Wallis}, \bibinfo{person}{Zeyuan Allen-Zhu},
  \bibinfo{person}{Yuanzhi Li}, \bibinfo{person}{Shean Wang},
  \bibinfo{person}{Lu Wang}, {and} \bibinfo{person}{Weizhu Chen}.}
  \bibinfo{year}{2021}\natexlab{}.
\newblock \showarticletitle{Lora: Low-rank adaptation of large language
  models}. In \bibinfo{booktitle}{\emph{arXiv preprint arXiv:2106.09685}}.
\newblock


\bibitem[Hu et~al\mbox{.}(2020)]%
        {hu2020open}
\bibfield{author}{\bibinfo{person}{Weihua Hu}, \bibinfo{person}{Matthias Fey},
  \bibinfo{person}{Marinka Zitnik}, \bibinfo{person}{Yuxiao Dong},
  \bibinfo{person}{Hongyu Ren}, \bibinfo{person}{Bowen Liu},
  \bibinfo{person}{Michele Catasta}, {and} \bibinfo{person}{Jure Leskovec}.}
  \bibinfo{year}{2020}\natexlab{}.
\newblock \showarticletitle{Open graph benchmark: datasets for machine learning
  on graphs}. In \bibinfo{booktitle}{\emph{NeurIPS'20}},
  Vol.~\bibinfo{volume}{33}. \bibinfo{pages}{22118--22133}.
\newblock


\bibitem[Jiang et~al\mbox{.}(2022)]%
        {jiang2022promptbert}
\bibfield{author}{\bibinfo{person}{Ting Jiang}, \bibinfo{person}{Jian Jiao},
  \bibinfo{person}{Shaohan Huang}, \bibinfo{person}{Zihan Zhang},
  \bibinfo{person}{Deqing Wang}, \bibinfo{person}{Fuzhen Zhuang},
  \bibinfo{person}{Furu Wei}, \bibinfo{person}{Haizhen Huang},
  \bibinfo{person}{Denvy Deng}, {and} \bibinfo{person}{Qi Zhang}.}
  \bibinfo{year}{2022}\natexlab{}.
\newblock \showarticletitle{Promptbert: improving bert sentence embeddings with
  prompts}. In \bibinfo{booktitle}{\emph{arXiv preprint arXiv:2201.04337}}.
\newblock


\bibitem[Jin et~al\mbox{.}(2023)]%
        {jin2023patton}
\bibfield{author}{\bibinfo{person}{Bowen Jin}, \bibinfo{person}{Wentao Zhang},
  \bibinfo{person}{Yu Zhang}, \bibinfo{person}{Yu Meng},
  \bibinfo{person}{Xinyang Zhang}, \bibinfo{person}{Qi Zhu}, {and}
  \bibinfo{person}{Jiawei Han}.} \bibinfo{year}{2023}\natexlab{}.
\newblock \showarticletitle{Patton: language model pretraining on text-Rich
  networks}. In \bibinfo{booktitle}{\emph{arXiv preprint arXiv:2305.12268}}.
\newblock


\bibitem[Kim et~al\mbox{.}(2020)]%
        {kim2020multimodal}
\bibfield{author}{\bibinfo{person}{Seungbae Kim}, \bibinfo{person}{Jyun-Yu
  Jiang}, \bibinfo{person}{Masaki Nakada}, \bibinfo{person}{Jinyoung Han},
  {and} \bibinfo{person}{Wei Wang}.} \bibinfo{year}{2020}\natexlab{}.
\newblock \showarticletitle{Multimodal post attentive profiling for influencer
  marketing}. In \bibinfo{booktitle}{\emph{WWW'20}}.
  \bibinfo{pages}{2878--2884}.
\newblock


\bibitem[Lester et~al\mbox{.}(2021)]%
        {lester2021power}
\bibfield{author}{\bibinfo{person}{Brian Lester}, \bibinfo{person}{Rami
  Al-Rfou}, {and} \bibinfo{person}{Noah Constant}.}
  \bibinfo{year}{2021}\natexlab{}.
\newblock \showarticletitle{The power of scale for parameter-efficient prompt
  tuning}. In \bibinfo{booktitle}{\emph{arXiv preprint arXiv:2104.08691}}.
\newblock


\bibitem[Li et~al\mbox{.}(2021b)]%
        {li2021adsgnn}
\bibfield{author}{\bibinfo{person}{Chaozhuo Li}, \bibinfo{person}{Bochen Pang},
  \bibinfo{person}{Yuming Liu}, \bibinfo{person}{Hao Sun},
  \bibinfo{person}{Zheng Liu}, \bibinfo{person}{Xing Xie},
  \bibinfo{person}{Tianqi Yang}, \bibinfo{person}{Yanling Cui},
  \bibinfo{person}{Liangjie Zhang}, {and} \bibinfo{person}{Qi Zhang}.}
  \bibinfo{year}{2021}\natexlab{b}.
\newblock \showarticletitle{Adsgnn: behavior-graph augmented relevance modeling
  in sponsored search}. In \bibinfo{booktitle}{\emph{SIGIR'21}}.
  \bibinfo{pages}{223--232}.
\newblock


\bibitem[Li et~al\mbox{.}(2021a)]%
        {li2021training}
\bibfield{author}{\bibinfo{person}{Guohao Li}, \bibinfo{person}{Matthias
  M{\"u}ller}, \bibinfo{person}{Bernard Ghanem}, {and} \bibinfo{person}{Vladlen
  Koltun}.} \bibinfo{year}{2021}\natexlab{a}.
\newblock \showarticletitle{Training graph neural networks with 1000 layers}.
  In \bibinfo{booktitle}{\emph{ICML'21}}. \bibinfo{pages}{6437--6449}.
\newblock


\bibitem[Li and Liang(2021)]%
        {li2021prefix}
\bibfield{author}{\bibinfo{person}{Xiang~Lisa Li} {and} \bibinfo{person}{Percy
  Liang}.} \bibinfo{year}{2021}\natexlab{}.
\newblock \showarticletitle{Prefix-tuning: optimizing continuous prompts for
  generation}. In \bibinfo{booktitle}{\emph{arXiv preprint arXiv:2101.00190}}.
\newblock


\bibitem[Liu et~al\mbox{.}(2021)]%
        {liu2021p}
\bibfield{author}{\bibinfo{person}{Xiao Liu}, \bibinfo{person}{Kaixuan Ji},
  \bibinfo{person}{Yicheng Fu}, \bibinfo{person}{Weng~Lam Tam},
  \bibinfo{person}{Zhengxiao Du}, \bibinfo{person}{Zhilin Yang}, {and}
  \bibinfo{person}{Jie Tang}.} \bibinfo{year}{2021}\natexlab{}.
\newblock \showarticletitle{P-tuning v2: prompt tuning can be comparable to
  fine-tuning universally across scales and tasks}. In
  \bibinfo{booktitle}{\emph{arXiv preprint arXiv:2110.07602}}.
\newblock


\bibitem[Liu et~al\mbox{.}(2019a)]%
        {liu2019roberta}
\bibfield{author}{\bibinfo{person}{Yinhan Liu}, \bibinfo{person}{Myle Ott},
  \bibinfo{person}{Naman Goyal}, \bibinfo{person}{Jingfei Du},
  \bibinfo{person}{Mandar Joshi}, \bibinfo{person}{Danqi Chen},
  \bibinfo{person}{Omer Levy}, \bibinfo{person}{Mike Lewis},
  \bibinfo{person}{Luke Zettlemoyer}, {and} \bibinfo{person}{Veselin
  Stoyanov}.} \bibinfo{year}{2019}\natexlab{a}.
\newblock \showarticletitle{Roberta: A robustly optimized bert pretraining
  approach}. In \bibinfo{booktitle}{\emph{arXiv preprint arXiv:1907.11692}}.
\newblock


\bibitem[Liu et~al\mbox{.}(2019b)]%
        {liu2019fine}
\bibfield{author}{\bibinfo{person}{Zhenghao Liu}, \bibinfo{person}{Chenyan
  Xiong}, \bibinfo{person}{Maosong Sun}, {and} \bibinfo{person}{Zhiyuan Liu}.}
  \bibinfo{year}{2019}\natexlab{b}.
\newblock \showarticletitle{Fine-grained fact verification with kernel graph
  attention network}. In \bibinfo{booktitle}{\emph{arXiv preprint
  arXiv:1910.09796}}.
\newblock


\bibitem[Mavromatis et~al\mbox{.}(2023)]%
        {mavromatis2023train}
\bibfield{author}{\bibinfo{person}{Costas Mavromatis},
  \bibinfo{person}{Vassilis~N Ioannidis}, \bibinfo{person}{Shen Wang},
  \bibinfo{person}{Da Zheng}, \bibinfo{person}{Soji Adeshina},
  \bibinfo{person}{Jun Ma}, \bibinfo{person}{Han Zhao},
  \bibinfo{person}{Christos Faloutsos}, {and} \bibinfo{person}{George
  Karypis}.} \bibinfo{year}{2023}\natexlab{}.
\newblock \showarticletitle{Train your own GNN teacher: graph-aware
  distillation on textual graphs}. In \bibinfo{booktitle}{\emph{arXiv preprint
  arXiv:2304.10668}}.
\newblock


\bibitem[Newman et~al\mbox{.}(2002)]%
        {newman2002random}
\bibfield{author}{\bibinfo{person}{Mark~EJ Newman}, \bibinfo{person}{Duncan~J
  Watts}, {and} \bibinfo{person}{Steven~H Strogatz}.}
  \bibinfo{year}{2002}\natexlab{}.
\newblock \showarticletitle{Random graph models of social networks}. In
  \bibinfo{booktitle}{\emph{Proceedings of the national academy of Sciences}},
  Vol.~\bibinfo{volume}{99}. \bibinfo{pages}{2566--2572}.
\newblock


\bibitem[Radford et~al\mbox{.}(2018)]%
        {radford2018improving}
\bibfield{author}{\bibinfo{person}{Alec Radford}, \bibinfo{person}{Karthik
  Narasimhan}, \bibinfo{person}{Tim Salimans}, \bibinfo{person}{Ilya
  Sutskever}, {et~al\mbox{.}}} \bibinfo{year}{2018}\natexlab{}.
\newblock \showarticletitle{Improving language understanding by generative
  pre-training}. In \bibinfo{booktitle}{\emph{OpenAI}}.
\newblock


\bibitem[Radford et~al\mbox{.}(2019)]%
        {radford2019language}
\bibfield{author}{\bibinfo{person}{Alec Radford}, \bibinfo{person}{Jeffrey Wu},
  \bibinfo{person}{Rewon Child}, \bibinfo{person}{David Luan},
  \bibinfo{person}{Dario Amodei}, \bibinfo{person}{Ilya Sutskever},
  {et~al\mbox{.}}} \bibinfo{year}{2019}\natexlab{}.
\newblock \showarticletitle{Language models are unsupervised multitask
  learners}. In \bibinfo{booktitle}{\emph{OpenAI}}.
\newblock


\bibitem[Reimers and Gurevych(2019)]%
        {reimers2019sentence}
\bibfield{author}{\bibinfo{person}{Nils Reimers} {and} \bibinfo{person}{Iryna
  Gurevych}.} \bibinfo{year}{2019}\natexlab{}.
\newblock \showarticletitle{Sentence-bert: sentence embeddings using siamese
  bert-networks}. In \bibinfo{booktitle}{\emph{arXiv preprint
  arXiv:1908.10084}}.
\newblock


\bibitem[Sun et~al\mbox{.}(2020)]%
        {sun2020ernie}
\bibfield{author}{\bibinfo{person}{Yu Sun}, \bibinfo{person}{Shuohuan Wang},
  \bibinfo{person}{Yukun Li}, \bibinfo{person}{Shikun Feng},
  \bibinfo{person}{Hao Tian}, \bibinfo{person}{Hua Wu}, {and}
  \bibinfo{person}{Haifeng Wang}.} \bibinfo{year}{2020}\natexlab{}.
\newblock \showarticletitle{Ernie 2.0: a continual pre-training framework for
  language understanding}. In \bibinfo{booktitle}{\emph{AAAI'20}},
  Vol.~\bibinfo{volume}{34}. \bibinfo{pages}{8968--8975}.
\newblock


\bibitem[Tian et~al\mbox{.}(2023)]%
        {tian2023graph}
\bibfield{author}{\bibinfo{person}{Yijun Tian}, \bibinfo{person}{Huan Song},
  \bibinfo{person}{Zichen Wang}, \bibinfo{person}{Haozhu Wang},
  \bibinfo{person}{Ziqing Hu}, \bibinfo{person}{Fang Wang},
  \bibinfo{person}{Nitesh~V Chawla}, {and} \bibinfo{person}{Panpan Xu}.}
  \bibinfo{year}{2023}\natexlab{}.
\newblock \showarticletitle{Graph neural prompting with large language models}.
  In \bibinfo{booktitle}{\emph{arXiv preprint arXiv:2309.15427}}.
\newblock


\bibitem[Touvron et~al\mbox{.}(2023)]%
        {touvron2023llama}
\bibfield{author}{\bibinfo{person}{Hugo Touvron}, \bibinfo{person}{Louis
  Martin}, \bibinfo{person}{Kevin Stone}, \bibinfo{person}{Peter Albert},
  \bibinfo{person}{Amjad Almahairi}, \bibinfo{person}{Yasmine Babaei},
  \bibinfo{person}{Nikolay Bashlykov}, \bibinfo{person}{Soumya Batra},
  \bibinfo{person}{Prajjwal Bhargava}, \bibinfo{person}{Shruti Bhosale},
  {et~al\mbox{.}}} \bibinfo{year}{2023}\natexlab{}.
\newblock \showarticletitle{Llama 2: Open foundation and fine-tuned chat
  models}. In \bibinfo{booktitle}{\emph{arXiv preprint arXiv:2307.09288}}.
\newblock


\bibitem[Vaswani et~al\mbox{.}(2017)]%
        {vaswani2017attention}
\bibfield{author}{\bibinfo{person}{Ashish Vaswani}, \bibinfo{person}{Noam
  Shazeer}, \bibinfo{person}{Niki Parmar}, \bibinfo{person}{Jakob Uszkoreit},
  \bibinfo{person}{Llion Jones}, \bibinfo{person}{Aidan~N Gomez},
  \bibinfo{person}{{\L}ukasz Kaiser}, {and} \bibinfo{person}{Illia
  Polosukhin}.} \bibinfo{year}{2017}\natexlab{}.
\newblock \showarticletitle{Attention is all you need}. In
  \bibinfo{booktitle}{\emph{NeurIPS'17}}, Vol.~\bibinfo{volume}{30}.
\newblock


\bibitem[Veli{\v{c}}kovi{\'c} et~al\mbox{.}(2017)]%
        {velivckovic2017graph}
\bibfield{author}{\bibinfo{person}{Petar Veli{\v{c}}kovi{\'c}},
  \bibinfo{person}{Guillem Cucurull}, \bibinfo{person}{Arantxa Casanova},
  \bibinfo{person}{Adriana Romero}, \bibinfo{person}{Pietro Lio}, {and}
  \bibinfo{person}{Yoshua Bengio}.} \bibinfo{year}{2017}\natexlab{}.
\newblock \showarticletitle{Graph attention networks}. In
  \bibinfo{booktitle}{\emph{arXiv preprint arXiv:1710.10903}}.
\newblock


\bibitem[Xu et~al\mbox{.}(2018)]%
        {xu2018powerful}
\bibfield{author}{\bibinfo{person}{Keyulu Xu}, \bibinfo{person}{Weihua Hu},
  \bibinfo{person}{Jure Leskovec}, {and} \bibinfo{person}{Stefanie Jegelka}.}
  \bibinfo{year}{2018}\natexlab{}.
\newblock \showarticletitle{How powerful are graph neural networks?}. In
  \bibinfo{booktitle}{\emph{arXiv preprint arXiv:1810.00826}}.
\newblock


\bibitem[Yang et~al\mbox{.}(2021)]%
        {yang2021graphformers}
\bibfield{author}{\bibinfo{person}{Junhan Yang}, \bibinfo{person}{Zheng Liu},
  \bibinfo{person}{Shitao Xiao}, \bibinfo{person}{Chaozhuo Li},
  \bibinfo{person}{Defu Lian}, \bibinfo{person}{Sanjay Agrawal},
  \bibinfo{person}{Amit Singh}, \bibinfo{person}{Guangzhong Sun}, {and}
  \bibinfo{person}{Xing Xie}.} \bibinfo{year}{2021}\natexlab{}.
\newblock \showarticletitle{GraphFormers: GNN-nested transformers for
  representation learning on textual graph}. In
  \bibinfo{booktitle}{\emph{NeurIPS'21}}, Vol.~\bibinfo{volume}{34}.
  \bibinfo{pages}{28798--28810}.
\newblock


\bibitem[Yang and Cui(2021)]%
        {yang2021bert}
\bibfield{author}{\bibinfo{person}{Yiping Yang} {and} \bibinfo{person}{Xiaohui
  Cui}.} \bibinfo{year}{2021}\natexlab{}.
\newblock \showarticletitle{Bert-enhanced text graph neural network for
  classification}. In \bibinfo{booktitle}{\emph{Entropy}},
  Vol.~\bibinfo{volume}{23}. \bibinfo{pages}{1--1}.
\newblock


\bibitem[Ying et~al\mbox{.}(2018)]%
        {ying2018graph}
\bibfield{author}{\bibinfo{person}{Rex Ying}, \bibinfo{person}{Ruining He},
  \bibinfo{person}{Kaifeng Chen}, \bibinfo{person}{Pong Eksombatchai},
  \bibinfo{person}{William~L Hamilton}, {and} \bibinfo{person}{Jure Leskovec}.}
  \bibinfo{year}{2018}\natexlab{}.
\newblock \showarticletitle{Graph convolutional neural networks for web-scale
  recommender systems}. In \bibinfo{booktitle}{\emph{KDD'18}}.
  \bibinfo{pages}{974--983}.
\newblock


\bibitem[Yuan and F{\"a}rber(2023)]%
        {yuan2023evaluating}
\bibfield{author}{\bibinfo{person}{Shuzhou Yuan} {and} \bibinfo{person}{Michael
  F{\"a}rber}.} \bibinfo{year}{2023}\natexlab{}.
\newblock \showarticletitle{Evaluating generative models for graph-to-text
  generation}. In \bibinfo{booktitle}{\emph{arXiv preprint arXiv:2307.14712}}.
\newblock


\bibitem[Zeng et~al\mbox{.}(2022)]%
        {zeng2022glm}
\bibfield{author}{\bibinfo{person}{Aohan Zeng}, \bibinfo{person}{Xiao Liu},
  \bibinfo{person}{Zhengxiao Du}, \bibinfo{person}{Zihan Wang},
  \bibinfo{person}{Hanyu Lai}, \bibinfo{person}{Ming Ding},
  \bibinfo{person}{Zhuoyi Yang}, \bibinfo{person}{Yifan Xu},
  \bibinfo{person}{Wendi Zheng}, \bibinfo{person}{Xiao Xia}, {et~al\mbox{.}}}
  \bibinfo{year}{2022}\natexlab{}.
\newblock \showarticletitle{Glm-130b: an open bilingual pre-trained model}. In
  \bibinfo{booktitle}{\emph{arXiv preprint arXiv:2210.02414}}.
\newblock


\bibitem[Zhao et~al\mbox{.}(2022)]%
        {zhao2022learning}
\bibfield{author}{\bibinfo{person}{Jianan Zhao}, \bibinfo{person}{Meng Qu},
  \bibinfo{person}{Chaozhuo Li}, \bibinfo{person}{Hao Yan},
  \bibinfo{person}{Qian Liu}, \bibinfo{person}{Rui Li}, \bibinfo{person}{Xing
  Xie}, {and} \bibinfo{person}{Jian Tang}.} \bibinfo{year}{2022}\natexlab{}.
\newblock \showarticletitle{Learning on Large-scale text-attributed graphs via
  variational inference}. In \bibinfo{booktitle}{\emph{arXiv preprint
  arXiv:2210.14709}}.
\newblock


\bibitem[Zhou et~al\mbox{.}(2019)]%
        {zhou2019semantic}
\bibfield{author}{\bibinfo{person}{Bolei Zhou}, \bibinfo{person}{Hang Zhao},
  \bibinfo{person}{Xavier Puig}, \bibinfo{person}{Tete Xiao},
  \bibinfo{person}{Sanja Fidler}, \bibinfo{person}{Adela Barriuso}, {and}
  \bibinfo{person}{Antonio Torralba}.} \bibinfo{year}{2019}\natexlab{}.
\newblock \showarticletitle{Semantic understanding of scenes through the ade20k
  dataset}. In \bibinfo{booktitle}{\emph{IJCV'19}}, Vol.~\bibinfo{volume}{127}.
  \bibinfo{pages}{302--321}.
\newblock


\end{thebibliography}
\appendix
\section{Experiment setting}
\label{appendix:data}
\begin{table}[h!]
	
	\caption{Statistics of experiment datasets.
	\normalsize
	}
	\label{tbl:exp:data}
 
\resizebox{0.48\textwidth}{!}{
		\begin{tabular}{c|c|c|c|c|c|c}
        \toprule
            Dataset & \# Nodes & \# Edges & \# Tokens &  Split ratio (\%) & \#Class&  Metric\\ \hline % & New-i(\%) & New-u(\%) \\ \hline
			Arxiv   & 169,343	& 1,166,243   & 35,920,710   & 54/18/28 & 40 & Accuracy \\ \hline %  & 59.5         & 96.1         \\ \hline
			Instagram  & 11,339      & 144,010   &  579,263   & 10/10/80 &2& ROC-AUC \\ \hline % & 58.1         & 97.4         \\ \hline
			Reddit  & 33,434      & 198,448   &  6,748,436 & 10/10/80 &2&Accuracy\\  %  & 23.7   
        \bottomrule

		\end{tabular}}
\end{table}
\subsection{Dataset Details}
We select three public and real-world datasets used for evaluation. Table \ref{tbl:exp:data} shows detailed statistics of these datasets. Below are the details of these datasets:\\ 
\noindent\textbf{Arxiv}. Ogbn-Arxiv (shorted as Arxiv), is a citation network where edges represent citation relationships, nodes represent papers and the text attribute is the abstracts of papers. The task on this graph is to predict paper subjects. This paper uses the public partition, ground truth, and text information provided by OGB\cite{hu2020open}. 

\noindent \textbf{Instagram}. Instagram is a social network where edges represent following relationships, nodes represent users, and the prediction task is to classify commercial and normal users in this network. The original dataset for Instagram is provided by \cite{kim2020multimodal}. Since the original dataset did not contain graph information, we obtained users' follow lists, personal introductions, and tags for commercial users through Instagram's public API \footnote{\url{https://developers.facebook.com/docs/graph-api}}. 

\noindent \textbf{Reddit}. Reddit is also a social network where each node denotes a user, the node features are the content of users' historically published subreddits, and edges denote whether two users have replied to each other. The prediction task is to classify whether a user is in the top 50\% popular (average score of all subreddits). It is constructed on a public dataset 
 \footnote{\url{https://convokit.cornell.edu/documentation/subreddit.html}} that collected replies and scores from Reddit users. The node text feature of this graph is the user's historical post content (limited to the last three posts per user). We divided users into popular and normal categories based on their average score of history posts, with users whose average score is higher than the median considered popular and others considered normal.
\subsection{Baselines}
\label{appendix:baselines}
We compare the proposed GraphAdapter with several state-of-the-art TAG modeling methods. 
\begin{itemize}[leftmargin=*]
\item \textbf{GNN-based methods}: This method directly combines different frozen PLM with GNNs to model TAGs. Since the specific GNN framework is not the key point this paper focuses on, this paper uses GraphSAGE \cite{hamilton2017inductive} as an instance of GNN.
%\DZ{Maybe 'this paper' is not good as 'this work'.}
\item \textbf{LM-based methods}: we select GIANT \cite{chien2021node}, and GLEM \cite{zhao2022learning} as baseline. GIANT uses self-supervised tasks to finetune PLM. Then incorporates the fine-tuned PLM and GNN to model TAG. GLEM jointly trains PLM and GNN. Note that GIANT is based on BERT, and GLEM uses DeBERTa. Considering PLMs have a high influence on performance, we also compare GraphAdapter with them under the same PLM.
\item \textbf{LLM-based methods}: There are a few LLM-based methods that are suitable in our setting. Therefore, we select TAPE \cite{he2023explanations} as the LLM-based baseline. This method, due to its need to obtain the interpretation of the text graph through GPT-3.5 and only the interpretation data on Arxiv is published. 
% However, Reddit and Instagram involve more than 40k nodes \DZ{in total}, and constructing their features is an expensive cost. 
Therefore, we only report the results of this method on Arxiv. Besides, we also extend MLPs, Graph2Text \cite{chen2023exploring}, and LoRA \cite{hu2021lora} to our experiment setting. Graph2Text directly incorporates the textual data of the 1-hop neighbors of a node to model the graph structure. For example, ``\textit{It is a paper node, its abstract is represented as XXX, and the abstracts of its cited papers are represented as follows: 1. YYY, 2. ZZZ.}''

\end{itemize}
Since many baseline methods involve GNN components, which are mostly optional, and considering that different GNNs have different performances. To make a fair comparison and without loss of generality, all GNNs used in all baselines are fixed to GraphSAGE, which is a classic and general GNN model. 

\subsection{Prompts}
\label{appendix:prompt}
According to the downstream task and graph information, this article has designed simple prompts for each dataset. As shown in Table \ref{tbl:promptsdetails}. It should be noted that because PLMs are sensitive to prompts, different prompts may result in significant performance differences. However, how to find suitable prompts is not the focus of this paper, so no search for prompts is conducted.

Most baselines rely on the sentence representations obtained from the LM. For instance, GNN+LM uses the sentence representation as a node feature. In baselines utilizing BERT or RoBERTa, we append the same prompts used in the original text to obtain prompt-aware sentence representations. When employing Llama 2, we use the same prompt and utilize the last token as the sentence representation.

\begin{table}[h]
  \caption{Detailed prompts on three datasets.}
    \label{tbl:promptsdetails}
  \centering
\resizebox{0.48\textwidth}{!}{
      \begin{tabular}{ccc}
        \toprule
        \textbf{Dataset}&\textbf{Node feature} &\textbf{prompts} \\ 
         \hline
        \textbf{Arxiv}& \{ABSTRACT\} & \makecell[l]{\textit{This is a paper's abstract:}\\ \{ABSTRACT\}.\\ \textit{Question: Based on the abstract above, this paper is published on} \\ \textit{\_\_\_ subject on Arxiv. Answer:} }\\
        \textbf{Instagram}& \{PROFILE\} &\makecell[l]{\textit{This is a user's profile is:}\\\{PROFILE\}\\\textit{Question: Based on the profile provided, this account is a} \\ \textit{\_\_\_ (answer in one word) account on Instagram. Answer:}}\\
        \textbf{Reddit}&\{LAST 3 POSTS \}&\makecell[l]{\textit{This is a user on Reddit, his last 3 posts are: }\\ \{LAST 3 POSTS\}. \\ \textit{Question: Based on the given posts, the style of this user is } \\ \textit{\_\_\_ (answer in one word). Answer:}}\\
        %\cmidrule(r){1-5}
        \bottomrule
      \end{tabular}
      }
\end{table}

\subsection{Implementation Details}
\label{appendix:implementation}
We independently pre-trained GraphAdapter on three datasets. The GNN used in the pre-training process was a 2-layer GraphSAGE, and the fusion layer used a 2-layer MLP. The pre-training was conducted for 50 rounds, and we used language model techniques such as silt activation function, layer-norm, and warm-up. The hidden side of GNN in GraphAdapter is set to 128, 64, and 128 on Arxiv, Instagram, and Reddit specifically.

When using BERT or RoBERTa with GraphAdapter, there are some modifications to the GraphAdapter pipeline. Since these language models (LMs) utilize a mask-prediction task, we modify the input of $\mathcal{S}_{i,k-1} = \{s{i,0}...s_{i,k-1},[mask],s_{i,k+1}...\}$ in Equation \ref{model:input}. Additionally, unlike based auto-regressive models, which use all tokens in pretraining, GraphAdapter based on BERT and RoBERTa only mask 20\% of tokens and pre-trained by their corresponding labels.

\begin{table*}[ht!]
    \caption{Performance of GraphAdapter with various GNN blocks. Here we fix LM as Llama 2 13B.}
\resizebox{0.85\textwidth}{!}{
\begin{tabular}{c|cc|cc|cc}
    \toprule
 & \multicolumn{2}{c|}{\textbf{Arxiv}} & \multicolumn{2}{c|}{\textbf{Instagram}} & \multicolumn{2}{c}{\textbf{Reddit}} \\
 GNN Block&GAT* & SAGE  &GAT* & SAGE &GAT* & SAGE \\
\midrule

Only GNN	           & 0.7534 \tiny{(0.0016)} & 0.7305  \tiny{(0.0020)} & 0.6292  \tiny{(0.0055)}
                             & 0.6221 \tiny{(0.0112)} & 0.6495  \tiny{(0.0031)} & 0.6320 \tiny{(0.0041)} 
                              \\
\hline
GraphAdapter(w/o pre)	           & 0.7621 \tiny{(0.0012)} & 0.7648  \tiny{(0.0020)} & 0.6490  \tiny{(0.0045)}
                             & 0.6351 \tiny{(0.0077)} & 0.6505  \tiny{(0.0070)} & 0.6369 \tiny{(0.0025)} 
                              \\
\hline

GraphAdapter                 & \textbf{0.7663 \tiny{(0.0016)}}  & \textbf{0.7707  \tiny{(0.0015)}} & \textbf{0.6545 \tiny{(0.0034)}}
                             & \textbf{0.6513 \tiny{(0.0075)}} & \textbf{0.6694 \tiny{(0.0041)}} & \textbf{0.6461 \tiny{(0.0019)}}
                             
	\\ \bottomrule
    \end{tabular}
    }
\label{tbl:ablation_gnnblock}
\end{table*} 
\begin{table}
    \caption{Results of additional ablation studies on GraphAdapter. ``r Fusion'' indicates replacing the Fusion component with sum-pooling, while ``o'' means using only a specific component.
    }
    \begin{tabular}{c|c|c|c}
    \toprule
      & {\textbf{Arxiv}} & {\textbf{Instagram}} & {\textbf{Reddit}} \\
\midrule
r Fusion          & 0.7698 \tiny{(0.0024)} & 0.6450 \tiny{(0.0080)} & 0.6361 \tiny{(0.0028)}  \\
 \hline
o GNN      & 0.7335 \tiny{(0.0024)} & 0.6294 \tiny{(0.0038)} & 0.6410 \tiny{(0.0027)} \\
\hline
o LLM+Prompt   & 0.7618 \tiny{(0.0019)}  & 0.6346 \tiny{(0.0044)} & 0.6092 \tiny{(0.0026)}  \\
 \hline
GraphAdapter             & 0.7707 \tiny{(0.0015)} & 0.6513 \tiny{(0.0075)} & 0.6461 \tiny{(0.0019)} 
	\\ \bottomrule
    \end{tabular}
    \label{tbl:ablation_appendix}
\end{table} 
\begin{table}[ht!]
    \caption{Running time of different methods on Arxiv using one Nvidia A800 80GB. 
    \small
    Since different methods use different PLM, we also report the number of parameters for the PLM (decoded as ``\# para'') and the number of trainable parameters (``\# trainable'').
    \normalsize
    }
 	\normalsize
  
\resizebox{0.45\textwidth}{!}{
    \begin{tabular}{c|c|c|c}
    \toprule
      & {\textbf{GIANT}} & {\textbf{GLEM}} & {\textbf{GraphAdapter}} \\
\midrule

PLM          &BERT&DeBERTa-Large&Llama 2-13B  \\ \hline
\# para of PLM          &110M&139M&13B  \\ \hline
\# trainable in Pre        &110M&-& 3M  \\ \hline 
\# trainable in Fine        &0.7M & 139M &2M \\

 \hline
 \hline
Pre-process          &-&-&192 min  \\ \hline
Pre-training       & 341 min & -&312 min \\ \hline
Fine-tuning    & 1 min & 612 min & 1 min \\\hline
Total time costs        & 342 min & 612 min& 505 min\\
\bottomrule
    \end{tabular}
\label{tbl:efficent}}
\end{table} 
\section{Experiment Result}
\subsection{Ablation Studies}
\label{appendix:ablation}
We also conduct experiments that isolate and specifically compare the contributions of the base model and the fusion of the graph-language model, aiming to enhance the robustness of \modelname. As shown in Table \ref{tbl:ablation_appendix}, the ranking of contributions from GNNs and LLMs varies across datasets. However, fusing GNNs and LLMs can achieve better performance in most cases. GraphAdapter not only efficiently combines GNNs and LLMs but also enhances their performance through next-token prediction pre-training.

Additionally, we investigate the effect of different GNNs with the same LM. In this experiment, we fix the LM as Llama 2 13B and compare the performance of GraphAdapter with different GNN blocks, both in pre-training and fine-tuning. We discovered that the original attention mechanism in Graph Attention Networks (GAT) is not effective for the pre-training of GraphAdapter. However, we found that a dot-product-based mechanism [3] yielded positive results. Consequently, we utilized a modified version of GAT, denoted as GAT*, and proposed this result to inspire future works. As Table \ref{tbl:ablation_gnnblock} shows, the pre-training of GraphAdapter is also suitable for attention-based GNN.

\begin{table*}[]
    \caption{Performance of different models with different prompts on Arxiv. Here the adopted LM is Llama 2-13B.
    }
    \normalsize
  
\resizebox{0.85\textwidth}{!}{
    \begin{tabular}{c|c|c|c}
    \toprule
      & {\textbf{LLM+MLP}} & {\textbf{GraphAdapter (w/o Pre)}} & {\textbf{GraphAdapter}} \\
\midrule
\makecell[l]{``\textit{Question: Based on the abstract above, this paper is published on} \\ \textit{\_\_\_ subject on Arxiv. Answer:}''}     & 0.7541  \tiny{(0.0024)} & 0.7648 \tiny{(0.0020)} & 0.7707  \tiny{(0.0015)} \\
 \hline
\makecell[l]{``\textit{Question: Please predict the subject it belongs to based on }\\ \textit{the abstract of this paper, please answer directly. Answer: }''}   & 0.7522  \tiny{(0.0021)} & 0.7657 \tiny{(0.0019)} &0.7719 \tiny{(0.0025)} \\ \hline

\makecell[l]{``\textit{Question: This paper is \_\_. Answer: }''}   & 0.7381 \tiny{(0.0021)} & 0.7554 \tiny{(0.0026)} & 0.7581 \tiny{(0.0028)} \\ \hline

\makecell[l]{``\textit{Question: I like \_\_ apple. Answer: }''}   & 0.7314 \tiny{(0.0010)} & 0.7478 \tiny{(0.0017)} & 0.7559 \tiny{(0.0017)} \\ \hline

\makecell[l]{None}   & 0.7335 \tiny{(0.0030)} & 0.7561 \tiny{(0.0014)} & 0.7607 \tiny{(0.0039)} \\

\bottomrule
    \end{tabular}
\label{tbl:prompt_arxiv}}
\end{table*}

\begin{table*}[]
    \caption{Performance of different models with different prompts on Instagram. Here the adopted LM is Llama 2-13B.
    }
    \normalsize
  
\resizebox{0.85\textwidth}{!}{
    \begin{tabular}{c|c|c|c}
    \toprule
      & {\textbf{LLM+MLP}} & {\textbf{GraphAdapter (w/o Pre)}} & {\textbf{GraphAdapter}} \\
\midrule
\makecell[l]{``\textit{Question: Based on the profile provided, this account is a} \\ \textit{\_\_\_ (answer in one word) account on Instagram. Answer:}''}     & 0.6248      \tiny{(0.0111)} & 0.6351 \tiny{(0.0077)} & 0.6513  \tiny{(0.0075)} \\
 \hline
\makecell[l]{``\textit{Based on the profile provided, please answer the type of} \\ \textit{this account(answer in one word). Answer:}''}   & 0.6325  \tiny{(0.0098)} & 0.6427 \tiny{(0.0044)} &0.6562 \tiny{(0.0031)} \\ \hline

\makecell[l]{``\textit{Question: This account is a \_\_. Answer: }''}   & 0.6298 \tiny{(0.0097)} & 0.6388 \tiny{(0.0100)} & 0.6392 \tiny{(0.0065)} \\ \hline

\makecell[l]{``\textit{Question: This user likes \_\_ apple. Answer: }''}   & 0.6161 \tiny{(0.0083)} & 0.6299 \tiny{(0.0107)} & 0.6308 \tiny{(0.0047)} \\ \hline

\makecell[l]{None}   & 0.6203 \tiny{(0.0071)} & 0.6306 \tiny{(0.0052)} & 0.6378 \tiny{(0.0055)} \\ 

\bottomrule
    \end{tabular}
\label{tbl:prompt_ins}}
\end{table*}

\begin{table*}[]
    \caption{Performance of different models with different prompts on Reddit. Here the adopted LM is Llama 2-13B.
    }
  
\resizebox{0.85\textwidth}{!}{
    \begin{tabular}{c|c|c|c}
    \toprule
      & {\textbf{LLM+MLP}} & {\textbf{GraphAdapter (w/o Pre)}} & {\textbf{GraphAdapter}} \\
\midrule
\makecell[l]{``\textit{Question: Based on the given posts, the style of this user is } \\ \textit{\_\_\_ (answer in one word). Answer:}''}     & 0.6123      \tiny{(0.0034)} & 0.6369 \tiny{(0.0025)} & 0.6461  \tiny{(0.0019)} \\
 \hline
\makecell[l]{``\textit{Based on the given posts, please answer the } \\ \textit{popularity of this user. Answer: }''}   & 0.6019  \tiny{(0.0021)} & 0.6324 \tiny{(0.0033)} &0.6380 \tiny{(0.0031)} \\ \hline

\makecell[l]{``\textit{Question: This user is \_\_. Answer: }''}   & 0.6117 \tiny{(0.0032)} & 0.6377 \tiny{(0.0022)} & 0.6446 \tiny{(0.0021)} \\ \hline

\makecell[l]{``\textit{Question: This user likes \_\_ apple. Answer: }''}   & 0.6103 \tiny{(0.0055)} & 0.6359 \tiny{(0.0044)} & 0.6413 \tiny{(0.0021)} \\ \hline

\makecell[l]{None}   & 0.6201 \tiny{(0.0020)} & 0.6354 \tiny{(0.0014)} & 0.6420 \tiny{(0.0024)} \\ 

\bottomrule
    \end{tabular}
\label{tbl:prompt_reddit}}
\end{table*}

\begin{table*}[ht]
    \caption{Three cases from the Ogbn-Arxiv dataset. LLM + MLP only utilizes abstract to predict paper's subjection, and make a wrong prediction on Case A and Case B. GraphAdapter (w/o Pre) can utilize both graph information and textual data, but also make a wrong prediction on Case B and Case C. After pretraining, GraphAdapter can make an accurate prediction on all cases.
    } 
\resizebox{0.85\textwidth}{!}{
    \begin{tabular}{c|c|c|c|c}
    \toprule
     & & {\textbf{Case A}} & {\textbf{Case B}} & {\textbf{Case C}} \\
\midrule
 \multirow{3}{*}{Feature}& Title & \makecell[l]{Text classification with \\pixel embedding} & \makecell[l]{Informative Image Captioning \\with External Sources of \\Information} & \makecell[l]{A Re-evaluation of Knowledge\\ Graph Completion Methods} \\ \cline{2-5}
& Abstract        & \makecell[l]{ Mentioned \\3x"convolutional",\\ 5x"kernel" 5X”3D”, \\but only \\2x"Text classification".}  & \makecell[l]{Focus on ``image caption''}&  \makecell[l]{Focus on ``Knowledge \\Graph Completion''}\\ \cline{2-5}

& Citation          & \makecell[l]{cited many \\ NLP papers.} & \makecell[l]{cited 5+ papers \\ from AAAI, and 5 papers \\ about ``Language'' }& \makecell[l]{cited many ``Machine Learning''\\ and ``Computation and \\Language'' papers.}\\
\hline
\multirow{3}{*}{Predict}&LLM+MLP  & Computer Vision & Computation and Language & Information Retrieval\\ \cline{2-5}
 & GraphAdapter (w/o Pre)                 & Computation and Language &Computer Vision & Machine Learning
	\\ \cline{2-5}
& GraphAdapter                 & Computation and Language & Computation and Language& Computation and Language
	\\ \hline
  &\textbf{Ground truth} & Computation and Language & Computation and Language & Computation and Language\\
 \bottomrule
    \end{tabular}}
\label{tbl:case_study}
\end{table*} 

\subsection{Analysis of Prompt}
\label{appendix:analysis_prormpts}
We further investigated GraphAdapter's stability with regard to prompts. We observed the performance of GraphAdapter based on different prompts. As shown in Tables \ref{tbl:prompt_arxiv}, \ref{tbl:prompt_ins}, and \ref{tbl:prompt_reddit}, task-related prompts significantly enhance the capability of large language models (LLMs) to handle text-as-graphs (TAGs). Comparing GraphAdapter (w/o pre) with LLMs+MLPs, it is evident that graph information is beneficial for prompts in most cases.
Simultaneously, GraphAdapter achieves additional improvements on top of prompts, demonstrating that the pretraining of GraphAdapter indeed facilitates the integration of graph information and prompts. Therefore, GraphAdapter stands out as a stable method suitable for different prompts

\subsection{Efficient}
\label{appendix:efficent}
Comparing time complexities of different methods is challenging due to varying compatible base language models. Therefore, we estimate time complexities as follows: Inference time/space complexity for a single node for the language model is $T_{infer}$ and $J_{infer}$, and training time/space complexity is $T_{train}$ and $J_{train}$. Complexity for non-linear transformations of PLM representations is $T_{MLP}$ and $J_{MLP}$. GIANT's complexity is equivalent to the time required for fine-tuning the PLM, i.e., $O(N \times T_{train})$ for training and $O(N \times T_{infer})$ for inference. GLEM has a similar complexity to GIANT. Our approach involves a single inference pass of PLMs, after which all operations are independent of PLMs. GraphAdapter only uses the processed representation to train GNN, resulting in $O(|S_{all}| * T_{GNN})$ complexity. $|S_{all}|$ is the total number of training tokens in the TAGs. Hence, our total complexity is $O(N \times T_{infer} + |S_{all}| \times T_{GNN})$. Our primary advantage is independence from $T_{train}$, and $T_{infer}$ can be accelerated by many methods. Considering larger language models where $T_{train}$ >> $T_{GNN}$, our approach holds a significant advantage. In terms of space complexity, our approach doesn't demand loading language model parameters during training, resulting in $O(batchsize \times J_{MLP})$ for Graph compared to $O(batchsize \times J_{GNN})$ for methods involving fine-tuning. Generally, $J_{PLM}$ >> $J_{GNN}$, allowing GraphAdapter to accommodate larger batch sizes in memory-restricted GPU environments. We also report efficiency comparisons for reproducibility purposes in Table \ref{tbl:efficent}.

\subsection{Case Studies}
\label{appendix:case}
We presented three cases in Table \ref{tbl:case_study}, considering different scenarios: 1. Node text acting as a distractor, 2. Graph features acting as distractors, 3. Neither the graph nor the text providing a strong signal. We denote these situations as Case A, Case B, and Case C, respectively. As shown in the cases, GraphAdapter without pretraining is effective when the text acts as a distractor (Case A, correct), but it struggles when the graph features are distracting (Case B, incorrect) due to over-reliance on graph information. After pretraining, GraphAdapter can more flexibly combine graph structural features and text features, enabling it to make judgments based on either graph or text information. Moreover, it can even make correct judgments on some very rare samples where neither the graph nor the text exhibits strong features. This indicates that pretrained GraphAdapter effectively utilizes the potential correlations between these two types of information.

\end{document}